\documentclass{article}

\usepackage{arxiv}

\usepackage[utf8]{inputenc} 
\usepackage[T1]{fontenc}    
\usepackage{hyperref}       
\usepackage{url}            
\usepackage{booktabs}       
\usepackage{amsfonts}       
\usepackage{nicefrac}       
\usepackage{microtype}      
\usepackage{lipsum}
\include{pythonlisting}
\usepackage[numbers]{natbib}
\usepackage{graphicx}
\usepackage{hyperref} 
\usepackage{multirow}
\usepackage{caption}
\usepackage[labelformat=simple]{subcaption}
\usepackage{amsmath,bm}
\usepackage[ruled,vlined]{algorithm2e}
\usepackage{dblfloatfix}
\usepackage{color}

\title{Blind Image Deblurring based on Kernel Mixture\thanks{This work has been submitted to the IEEE for possible publication. Copyright may be transferred without notice, after which this version may no longer be accessible.}}

\author{
  Sajjad Amrollahi Biyouki \\
  Department of Industrial and Systems Engineering \\
  The University of Tennessee\\
  Knoxville, TN 37996 \\
  \texttt{samrolla@vols.utk.edu} \\
  
   \And
 Hoon Hwangbo \\
  Department of Industrial and Systems Engineering \\
  The University of Tennessee\\
  Knoxville, TN 37996 \\
  \texttt{hhwangb1@utk.edu} \\
}

\begin{document}
\maketitle

\begin{abstract}
Blind Image deblurring tries to estimate blurriness and a latent image out of a blurred image. This estimation, as being an ill-posed problem, requires imposing restrictions on the latent image or a blur kernel that represents blurriness. Different from recent studies that impose some priors on the latent image, this paper regulates the structure of the blur kernel. We propose a kernel mixture structure while using the Gaussian kernel as a base kernel. By combining multiple Gaussian kernels structurally enhanced in terms of scales and centers, the kernel mixture becomes capable of modeling nearly non-parametric shape of blurriness. A data-driven decision for the number of base kernels to combine makes the structure even more flexible. We apply this approach to a remote sensing problem to recover images from blurry images of satellite. This case study shows the superiority of the proposed method regulating the blur kernel in comparison with state-of-the-art methods that regulates the latent image.  
\end{abstract}


\section{Introduction}
Image restoration is widely used when images do not provide desired quality in terms of clarity and contrast of target objects as a consequence of noisy disturbance so-called blurriness. Such poor-quality images are often observed in remote sensing \cite{shen2012blind},\cite{zhang2019scale}, underwater objects detection \cite{li2016underwater},\cite{peng2017underwater},
and healthcare applications \cite{jiang2003blind},\cite{you2019ct}. Due to great need in many real-world applications, a wide spectrum of methods have been developed to restore a ``best-quality'' latent image from a blurred image \cite{fergus2006removing},\cite{cho2009fast},\cite{shan2008high}. 

In general, a blur process is modeled as the convolution of a latent image and blurriness represented by a kernel. The deconvolution method, the inverse process of the convolution, is the process of extracting the latent image by deconvolving a blurred image into the latent image and the blur kernel. If some prior knowledge is given for the blur kernel so if a specific kernel can be assumed \textit{a priori}, the deconvolution process becomes straightforward as it only requires estimating the latent image.  This type of approaches that use a pre-defined kernel are called non-blind deconvolution \cite{richardson1972bayesian},\cite{bar2006image}. The kernels are, however, unknown or only partially known in real-world problems. With the rapid increase in the usage of images for an analysis of various systems, assuming a pre-defined kernel is often too restrictive.  In this regard, the blind deconvolution  \cite{almeida2009blind},\cite{levin2011understanding} (also referred to as blind image deblurring) that estimates both the latent image and the blur kernel has attracted great attentions in the last two decades.  

The blind deconvolution process requires solving an ill-posed problem. This is because, for a given, single blurred image, there can be an infinite number of solutions for the latent image and the blur kernel that satisfy the system of equations defined for the blur process model:
\begin{equation}
    \mathbf{B} = \mathbf{I} * \mathbf{K} + \mathbf{n}
    \label{blur process}
\end{equation}
where $*$ is the convolution operator, $\mathbf{B}$ is a degraded (blurred) image, $\mathbf{I}$ is a recovered (latent) image, $\mathbf{K}$ is a blur kernel, and $\mathbf{n}$ is an additive noise. Therefore, to derive a unique solution for the blind deconvolution problem, additional restrictions need to be imposed.  For this purpose, some priors and regularization terms have been used to maintain and intensify image edges (restricting latent images) or to diminish harmful degradation and noise (restricting blur kernels).  Low-rank prior \cite{ren2016image}, dark channel prior \cite{pan2016blind}, and graph-based prior \cite{bai2018graph} have been developed recently to restrict the latent image, and total variation (TV) regularization \cite{ chan1998total} and gradient prior \cite{xu2010two} have been applied to either the blur kernel or the latent image.

Recently, the approaches restricting the latent image, e.g., through statistical priors \cite{fergus2006removing},
\cite{pan2016$l_0$} or image related priors \cite{ren2016image},\cite{pan2016blind},\cite{bai2018graph}, have been a mainstream in blind image deblurring research. However, image characteristics can vary by different images and applications, and there is no guarantee that the restrictions justified for a specific latent image work well for other blind deconvolution problems. On the other hand, blurriness is typically originated from common sources, including  atmospheric turbulence (mostly for remote sensing), out-of-focus, camera shake or object motion \cite{yan2016blind},\cite{lee2017mixture}.  This suggests that restricting the blur kernel rather than the latent image can be more effective and generalized more easily to broad applications of the blind deconvolution. In this paper, we propose a novel kernel structure to model the blur kernel, which can be leveraged for any blurry image no matter what image prior (on the latent image) is appropriate.

A Gaussian kernel has been used to model the blur kernel, especially to estimate atmospheric turbulence blurs \cite{yan2016blind}. In general, a circular shape modeled by a simple Gaussian blur kernel is not sufficient to represent the underlying blurriness, and there could be multiple sources of blurriness making the overall shape complicated. To address this problem, we propose using a mixture of several Gaussian kernels as a blur kernel while allowing different scales, centers, and rotations of each individual kernel. This kernel mixture is capable of modeling a complex shape of blurriness
from symmetric to asymmetric and from circular to linear without significant limitation. Consequently, the kernel mixture shows flexible behaviors as if it were estimated nonparametrically even with its parameterized structure. All the decisions to define the kernel structure, including how many base kernels to combine, are data-driven, so the proposed kernel mixture is adaptive and can be applied to any blind deconvolution problem.

The main contributions of this paper can be summarized as:
\begin{itemize}
    \item This paper develops a novel kernel structure, a kernel mixture, that can be applied to a broad class of blind image deblurring problems, independent of the characteristics of latent images.
    \item The parametric structure of the proposed kernel, induced by Gaussian base kernels, restricts and characterizes the blur kernel effectively producing a good solution for the ill-posed blind deconvolution problem.
    \item The proposed kernel is flexible in modeling blurriness; with different scales, centers, and rotations, the Gaussian kernels become capable of modeling various shapes of blurriness.
    \item The proposed kernel is adaptive to given images since the determination of its structure is data-driven.
\end{itemize}

The rest of the paper is organized as follows. Section~\ref{sc:lit rev}  reviews the related works in the blind image deblurring domain. Section~\ref{sc:method} describes the proposed method in detail elaborating the development of the kernel mixture, the optimization of associated parameters, and the overall process of blind deconvolution based on the kernel mixture. Section~\ref{sc:case} presents a case study of deblurring noisy satellite images, discusses the dataset and experimental settings, and compares the proposed method with other state-of-the-art benchmark methods. Section~\ref{sc:concl} discusses future research directions and concludes the paper.

\section{Related Works}\label{sc:lit rev}
Image deconvolution methods can be grouped into two general categories: the non-blind deconvolution where the kernel information is known and the blind deconvolution where the kernel is also unknown and subject to estimation.
In the non-blind deconvolution domain, Wiener filter \cite{wiener1950extrapolation} and Richardson-Lucy algorithm \cite{richardson1972bayesian} are the most well-known methods among earlier works, and they are still in use for the image restoration problems. The major shortcoming of these methods is in their noise sensitivity that leads to ringing artifacts in the recovered image. In addition, these methods require assuming a specific kernel, but it is hard to find a proper kernel that works well for different images/applications.
Albeit more difficult to implement, the blind deconvolution has been used more broadly with better capability of image recovery in general. Earlier studies mostly focused on removing motion blur \cite{cho2009fast},\cite{fergus2006removing},\cite{shan2008high} caused by dynamic movement of an object while an image is taken. 
Recent papers also considered other types of blurriness stemmed from various sources such as atmosphere turbulence and camera shake \cite{bai2018graph},
\cite{yan2016blind}.
To solve a blind deconvolution problem, some Bayesian approaches, specifically Maximum a Posteriori (MAP) estimation, and other optimization techniques have been used.

A few decades ago, \citet{likas2004variational} proposed using a hierarchical Bayesian modeling for blind deblurring. They used Gaussian probability distribution for modeling image prior, blur kernel, and hyperparameters of the priors. They employed the variational Expectation Maximization (EM) to obtain the MAP estimates from their Bayesian model. \citet{fergus2006removing} used Miskin's ensemble learning \cite{miskin2000ensemble} for a variational Bayesian approach while assuming a Gaussian mixture prior for the latent image. Inspired by this study, other researchers proposed more efficient approaches by considering various priors  \cite{babacan2012bayesian},\cite{babacan2008variational},\cite{molina2006blind}. 
\citet{babacan2012bayesian} employed the variational Bayesian approach while assuming sparse priors for the latent image (super-Gaussian priors).
\citet{babacan2008variational} imposed total variation prior on the latent image and assumed a Gaussian blur kernel. \citet{molina2006blind} proposed simultaneous autoregressions as priors for both the latent image and blur kernel and used gamma distributions to model the hyperparameters of the priors.  The major weakness of this type of approaches based on the MAP estimation is their strong dependency on the choice of priors and the lack of generality as a consequence. It has been shown that the blind deconvolution tends to estimate a trivial unblur image when an MAP approach is applied \cite{babacan2012bayesian},\cite{levin2011understanding}.  In addition, when sparse priors are used, the computational performance of an MAP estimation exacerbates as the objective function for the estimation becomes non-convex \cite{xu2010two}.

Others have used optimization techniques to solve a blind deconvolution problem. The main idea is to solve an individual optimization problem for each of the latent image and blur kernel while keeping one constant in the estimation of the other and iteratively updating the estimates \cite{you1996regularization}, \cite{chan1998total}. \citet{you1996regularization} introduced such an alternating optimization problem in which they regularized both the latent image and blur kernel by using the Laplace operator. \citet{chan1998total} took advantage of the alternating optimization structure and proposed the usage of total variation regularization, which can improve recovering the edges of an image. Since then, the alternating optimization-based approaches have been evolved in two different ways: i) introducing novel image priors and ii) developing new blur kernel structures. 

Most recent works have developed more sophisticated image priors, including \textit{$l_0$}-norm prior \cite{pan2016$l_0$}, low-rank prior \cite{ren2016image}, dark channel prior \cite{pan2016blind}, and most recently, reweighed graph-based prior \cite{bai2018graph}. \citet{pan2016$l_0$} proposed using the $l_0$-norm that regulates the number of nonzero pixels as it can distinguish a clear image from a blurred image based on their opposite behaviors in terms of nonzero intensities. Regulating the $l_0$-norm prior, however, makes the objective function for estimating the latent image and blur kernel non-convex, so the estimation problem becomes hard to solve \cite{bai2018graph}.
\citet{ren2016image} enforced image prior to be low-rank approximation of a degraded image by using a weighted nuclear norm minimization and combined the image prior with the gradient map of the image. However, this requires solving pixel-based singular value decomposition that has $O(N^3)$ complexity \cite{bai2018graph}. 
\citet{pan2016blind} observed that the dark channel of blurred images should be less sparse than that of clear images due to the nature of a typical blur process. Based on this observation, they proposed regulating the dark channel of the latent image and making it sparse. This process though involves computing nonlinear dark channel and its $l_0$-norm, so implementing the approach is computational intensive \cite{bai2018graph}. Most recently, \citet{bai2018graph} incorporated a graph structure to model a blurred image in nodal domain and used the graph structure as the image prior. This was the first effort mapping pixels to a domain other than frequency and real domains. All the discussed methods with novel image priors have improved the quality of image recovery.  However, they may not be effective in extracting a latent image if the characteristics assumed for the priors are not so obvious in a given image.  In other words, their quality can vary substantially depending on the images given.

In the domain of blur kernel regularization, \citet{xu2010two} utilized influential edges of an image (image gradients) to create an initial kernel and refined the kernel by using an iterative support detection. Although this approach regularizes the blur kernel, the kernel estimation strongly relies on the construction of edges of which characteristics vary by images. 
More recently, other studies have imposed specific structures on the blur kernel by combining multiple kernels of known structures.
\citet{mai2015kernel} fused blur kernels estimated from other methods by using Gaussian Conditional Random Fields. 
Their approach outperformed the other methods from which the individual kernels were extracted. 
However, this requires implementing all the other methods, and the entire estimation process becomes computationally expensive. Furthermore, its performance depends on the quality of individual kernels. If none of the individual kernels is capable of capturing specific properties of the true blurriness, the kernel fusion will also fail to model such a property.
Later, \citet{lee2017mixture} modeled a blur kernel as a linear combination of basic two-dimensional patterns. To construct the patterns, they used ``one-dimensional'' Gaussian density with a different scale parameter for each pattern. Since this blur kernel relies on simple Gaussian densities, it cannot represent various shapes of blurriness. In this paper, we develop a kernel mixture that combines structure-enhanced Gaussian kernels, which is capable of modeling almost any shape of blurriness. 

\section{Methodology}\label{sc:method}
In this section, we develop the kernel mixture of structure-enhanced Gaussian kernels and present its capability of modeling general shapes of blurriness.  We also propose a blind image deblurring formulation that properly regulates the latent image and blur kernel in the context of the kernel mixture.  Then, we describe the alternating optimization problem used to estimate the blur kernel and latent image. 

\subsection{Mixture of Structure-Enhanced Gaussian Kernels}
The major objective of this section is to develop a blur kernel that is flexible in modeling blurriness while maintaining a certain parametric structure. We use Gaussian kernels as base kernels to impose parametric structures and improve the structures in terms of scales, centers, and rotations to grant various shape characteristics when they are combined.

The general model of our proposed blur kernel $\mathbf{K}$ is defined as a mixture of $N$ Gaussian kernels as
\begin{equation}
\mathbf{K} = \sum_{t=1}^N K_{G,t}
\label{Kernel-sum}
\end{equation}
where $K_{G,t}$ for $t=1,\ldots,N$ is a base Gaussian kernel. The base kernels will have different shape complexity which is determined by the underlying blurriness of a given image. We present possible base kernel structures and the shape of their mixture from the simplest to the most complicated ones with their distinguished specifications.

\begin{figure}[h]
    \centering
    \includegraphics[width=.25\textwidth]{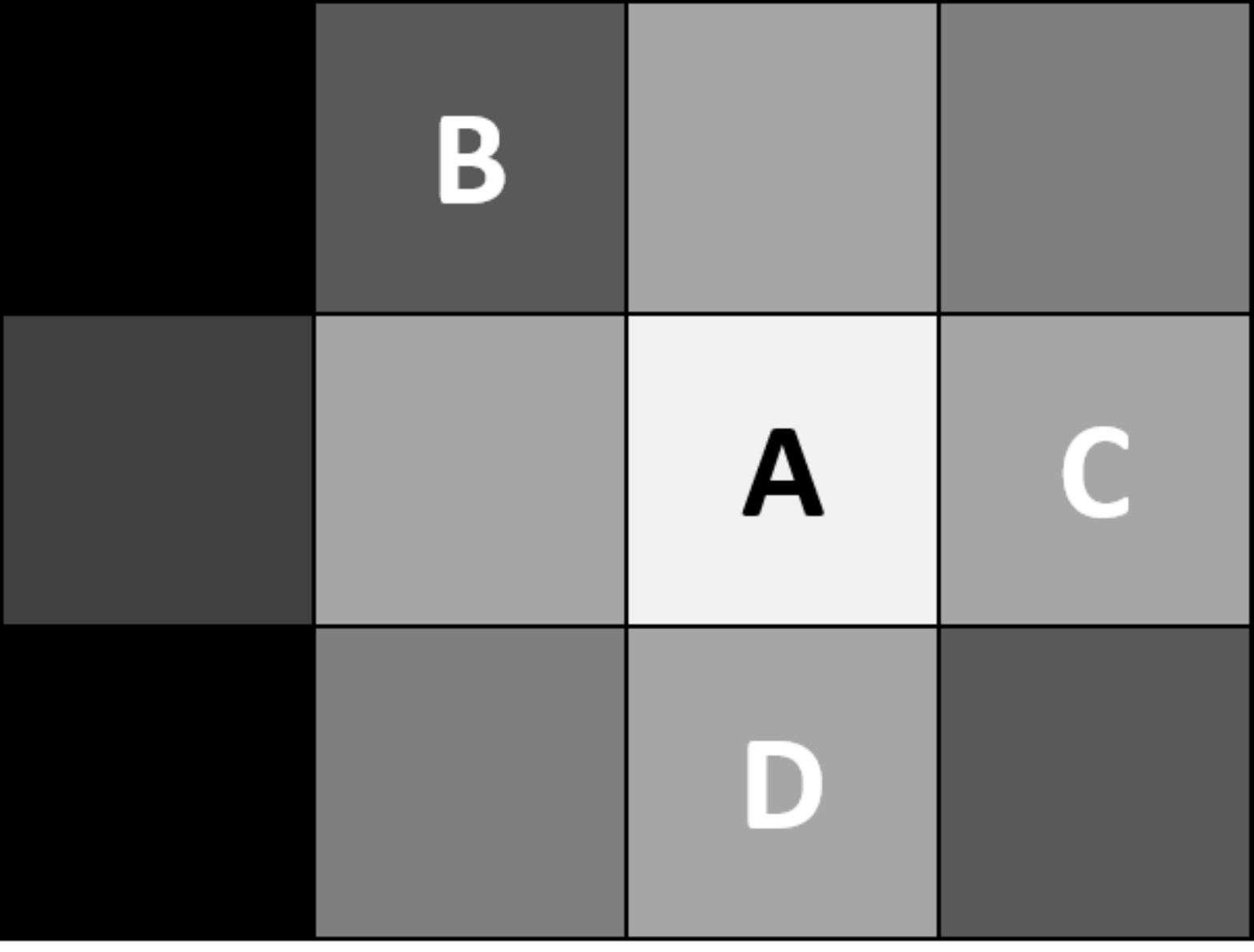}
    \caption{Relative locations of pixels}
    \label{fig:rel loc}
\end{figure}

The simplest structure of a two-dimensional base kernel is an isotropic zero-mean Gaussian kernel with a scale parameter $\sigma$:
\begin{equation}
    K_{G}^{\text{simple}}(x,y;\sigma^2) \propto \exp{(-\frac{x^2 + y^2}{2\sigma^2})}
    \label{basic_gaussian}
\end{equation}
where $x$ and $y$ are the locations of a pixel being evaluated, relative to a target pixel location in the horizontal and vertical axes of an image, respectively. For example, suppose pixel~A in Fig.~\ref{fig:rel loc} is a target pixel. When evaluating blurriness propagating from the target pixel to others, the location of B relative to A is calculated by $(x,y)=(-1,1)$ as it moves to the left and to the top each by one pixel.  Similarly, the location of C and D are $(1,0)$ and $(0,-1)$, respectively.

In Eq.~\eqref{basic_gaussian}, the kernel $K_{G}^{\text{simple}}$ is determined by the single scale parameter $\sigma$.  When using a single parameter to describe the scale toward all directions, the propagation of blurriness is modeled as a circular shape. Even the mixture of $K_{G}^{\text{simple}}$ will remain as a concentric circular shape. As such, this kernel is too simple to represent general blurriness.

\begin{figure}[b]
    \centering
    \begin{subfigure}[t]{3.5cm}
       \captionsetup{justification=centering,margin=0.1mm}
        \centering
    
        \includegraphics[width=3.5cm]{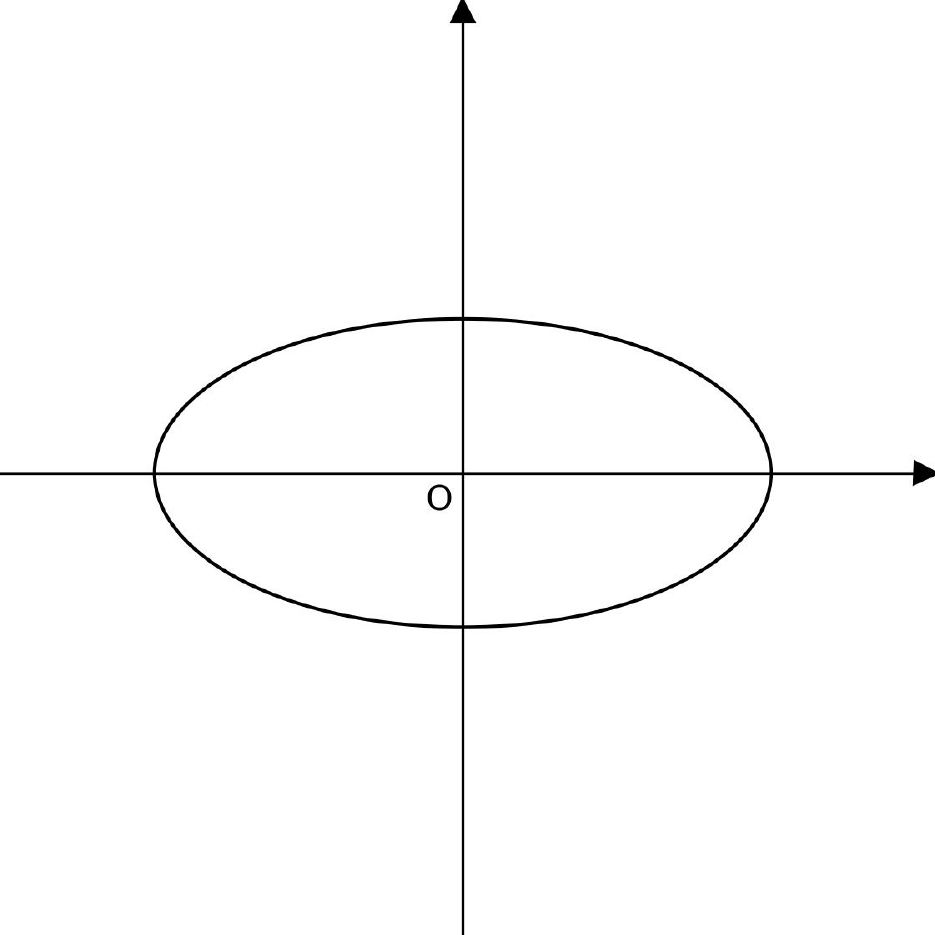}
        \caption{Blurriness shape with individual scales}
        \label{fg:shape scale}
    \end{subfigure}
    ~
     \begin{subfigure}[t]{3.5cm}
     \captionsetup{justification=centering,margin=0.1mm}
        \centering
        \includegraphics[width=3.5cm]{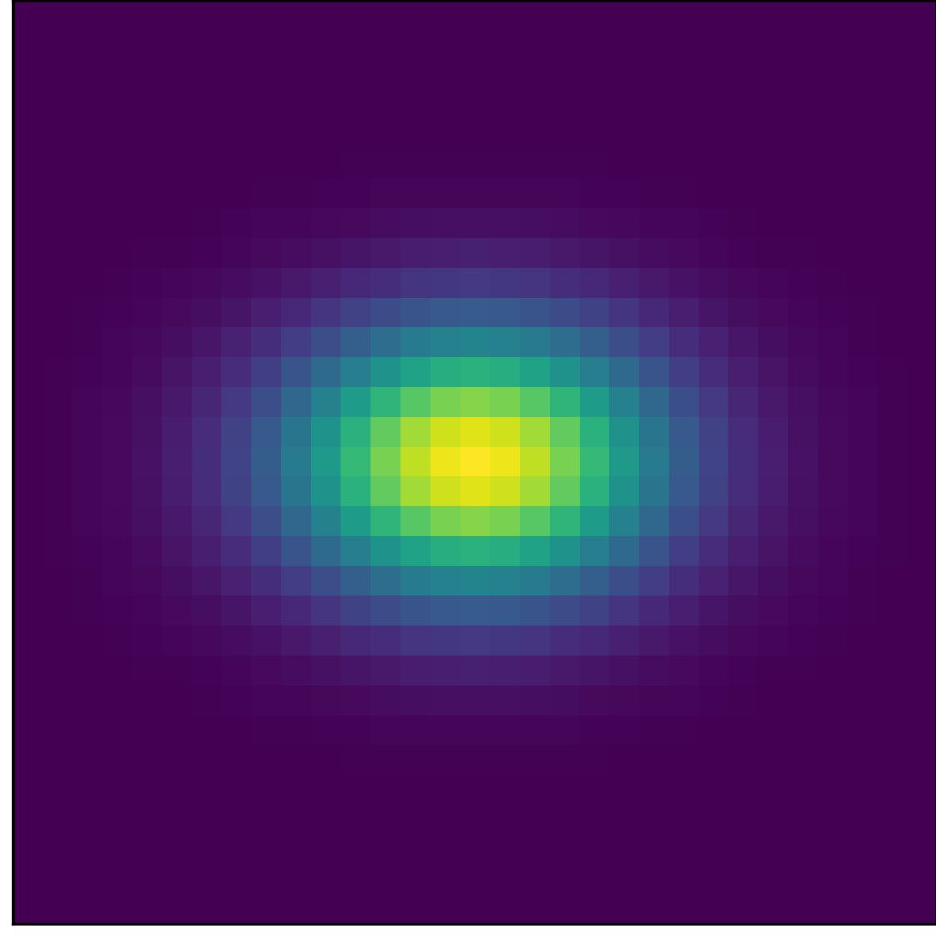}
        \caption{Scale-enhanced base kernel}
    \end{subfigure}
    ~
     \begin{subfigure}[t]{3.5cm}
        \captionsetup{justification=centering,margin=0.1mm}
        \centering
        \includegraphics[width=3.5cm]{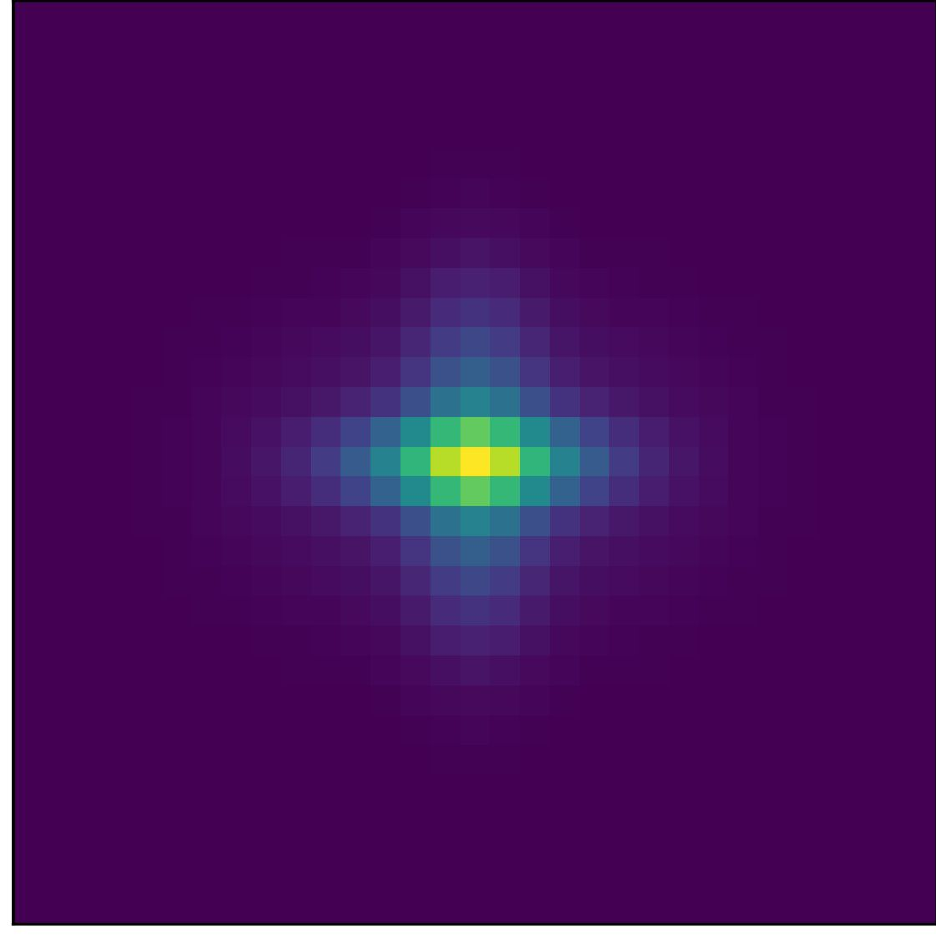}
        \caption{Mixture of $K_{G}^\text{scale}$}
        \label{cross lines}
    \end{subfigure}
    \caption{Kernel mixture with scale enhancement}
    \label{simple gaussian}
\end{figure}

To improve the simple structure, we allow a different scale parameter for each axis and construct a covariance matrix to account for the varying scales of different axes. A scale-enhanced Gaussian kernel is expressed in matrix form as
\begin{equation}
    K_{G}^\text{scale}(\mathbf{p};\bm{\Sigma}) \propto
    \exp{(-\frac{1}{2}
    \mathbf{p}^T
    \bm{\Sigma}^{-1}
    \mathbf{p})}
    \label{covariance gaussian}
\end{equation}
where $\mathbf{p} =  \begin{bmatrix}
    x & y
    \end{bmatrix}^T$, $\bm{\Sigma} = \text{diag}(\sigma_x^2, \sigma_y^2)$, and $\sigma_x^2$ and $\sigma_y^2$ are the variances associated with each axis.
By allowing different scales on each axis, this kernel can be represented as an elliptic shape as shown in Fig.~\ref{fg:shape scale}. In an extreme condition, as one of $\sigma_x^2$ or $\sigma_y^2$ gets close to zero, the elliptic shape can converge to a linear shape since the kernel is defined on the discretized grid formed by the pixels. A combination of $K_{G}^\text{scale}$ kernels produces a centralized multi-layer cross-line shape as shown in Fig.~\ref{cross lines}.

\begin{figure}[t]
    \centering
    \begin{subfigure}[t]{3.5cm}
        \centering
        \captionsetup{justification=centering}
        \includegraphics[width=3.5cm]{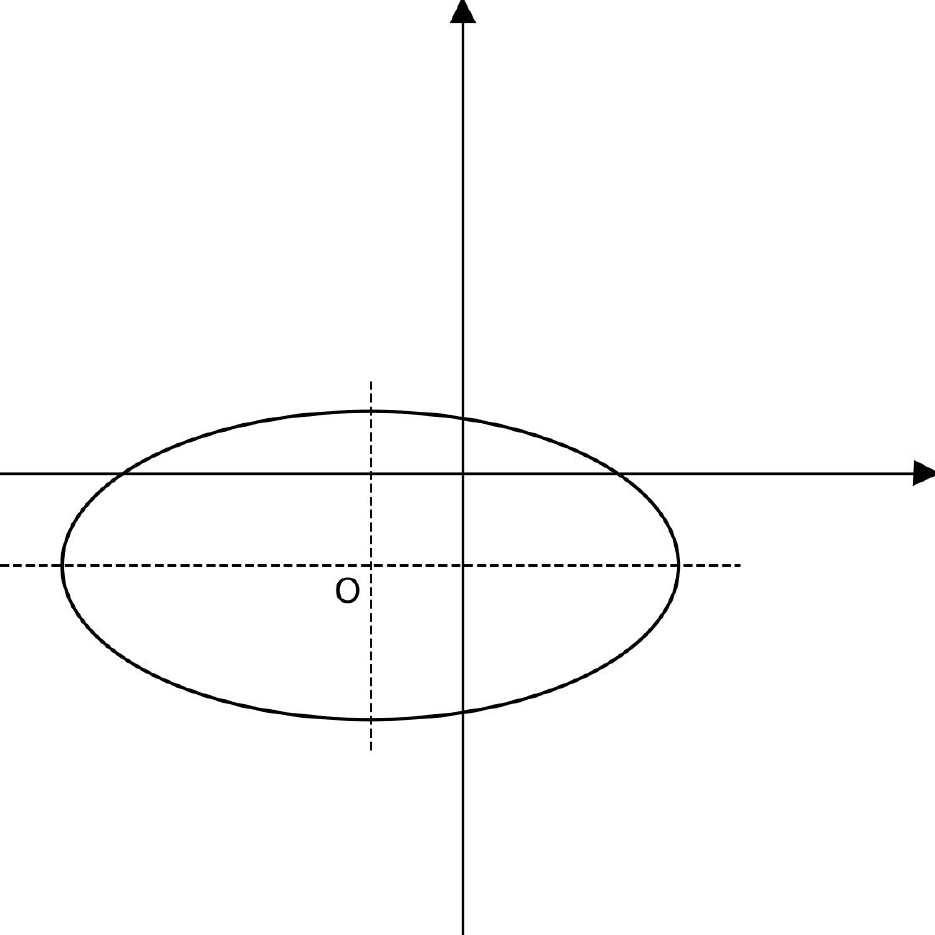}
        \caption{Blurriness shape with a non-zero center}
    \end{subfigure}
    ~
     \begin{subfigure}[t]{3.5cm}
        \centering
        \captionsetup{justification=centering}
        \includegraphics[width=3.5cm]{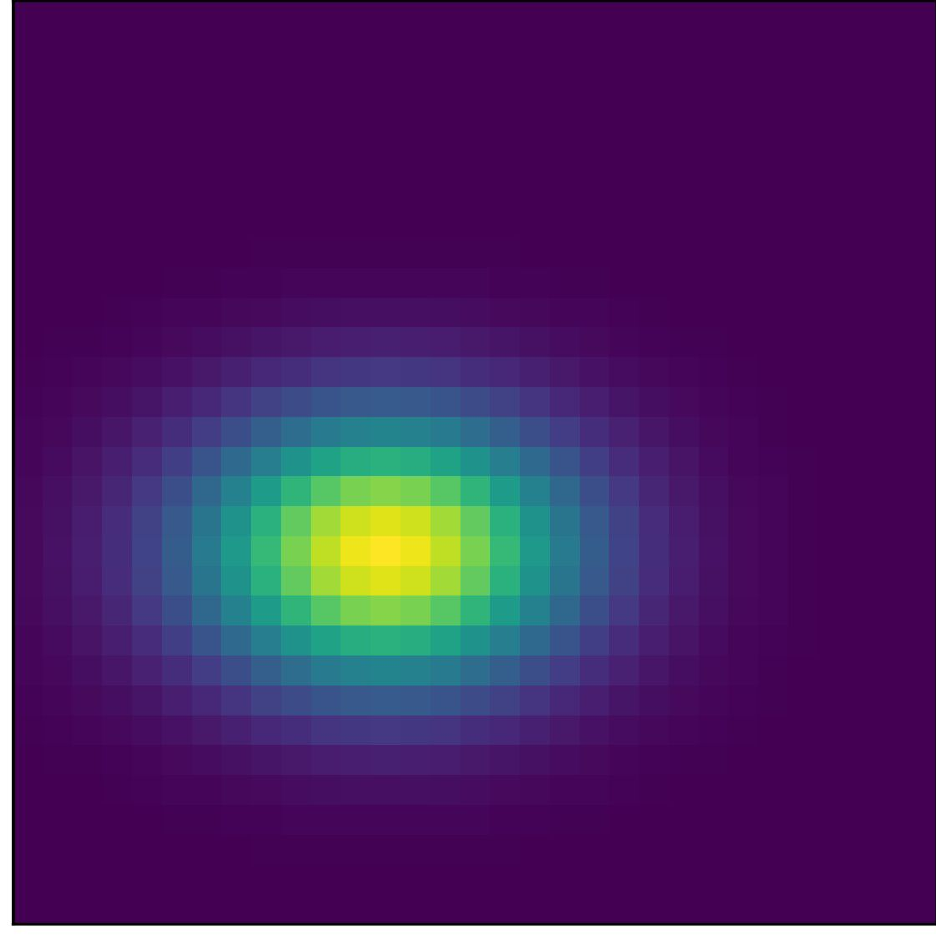}
        \caption{Center-enhanced base kernel}
    \end{subfigure}
    ~
     \begin{subfigure}[t]{3.5cm}
        \centering
        \captionsetup{justification=centering}
        \includegraphics[width=3.5cm]{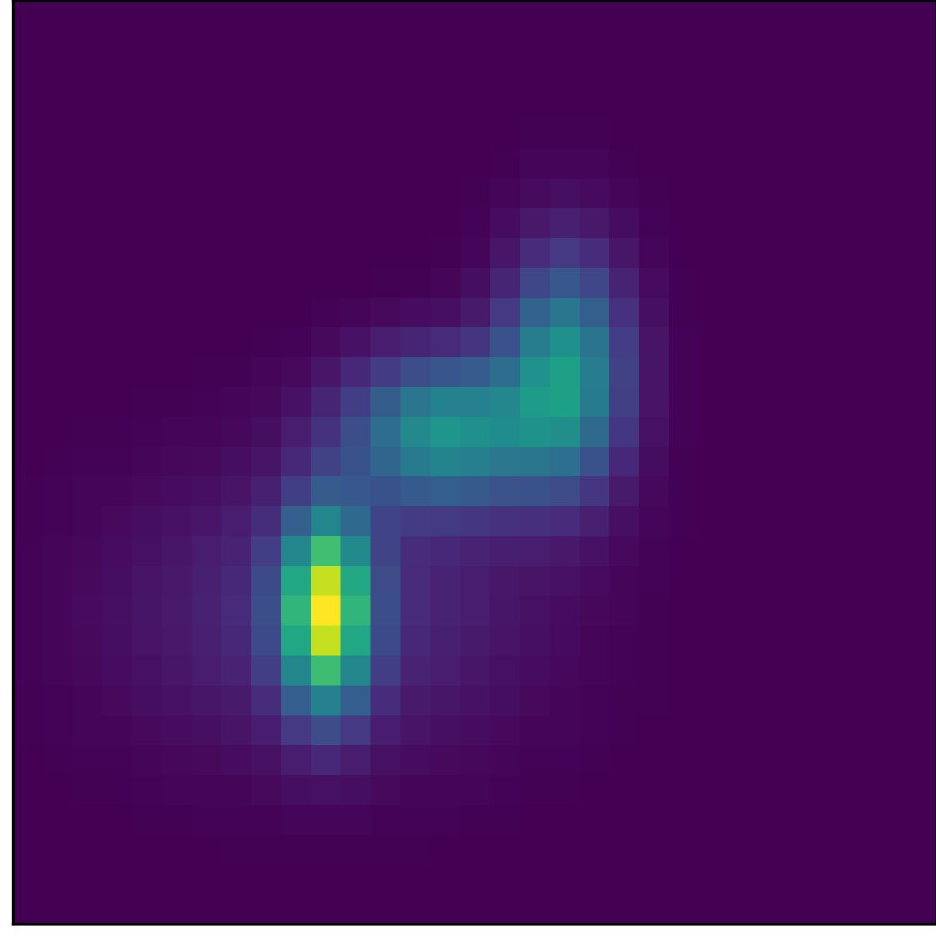}
        \caption{Mixture of $K_{G}^\text{center}$}
        \label{partc-nondefault}
    \end{subfigure}
    \caption{Kernel mixture with center enhancement}
    \label{Non-default gaussian}
\end{figure}
\begin{figure}[t]
\centering
    \begin{subfigure}[t]{3.5cm}
        \centering
        \captionsetup{justification=centering}
        \includegraphics[width=3.5cm]{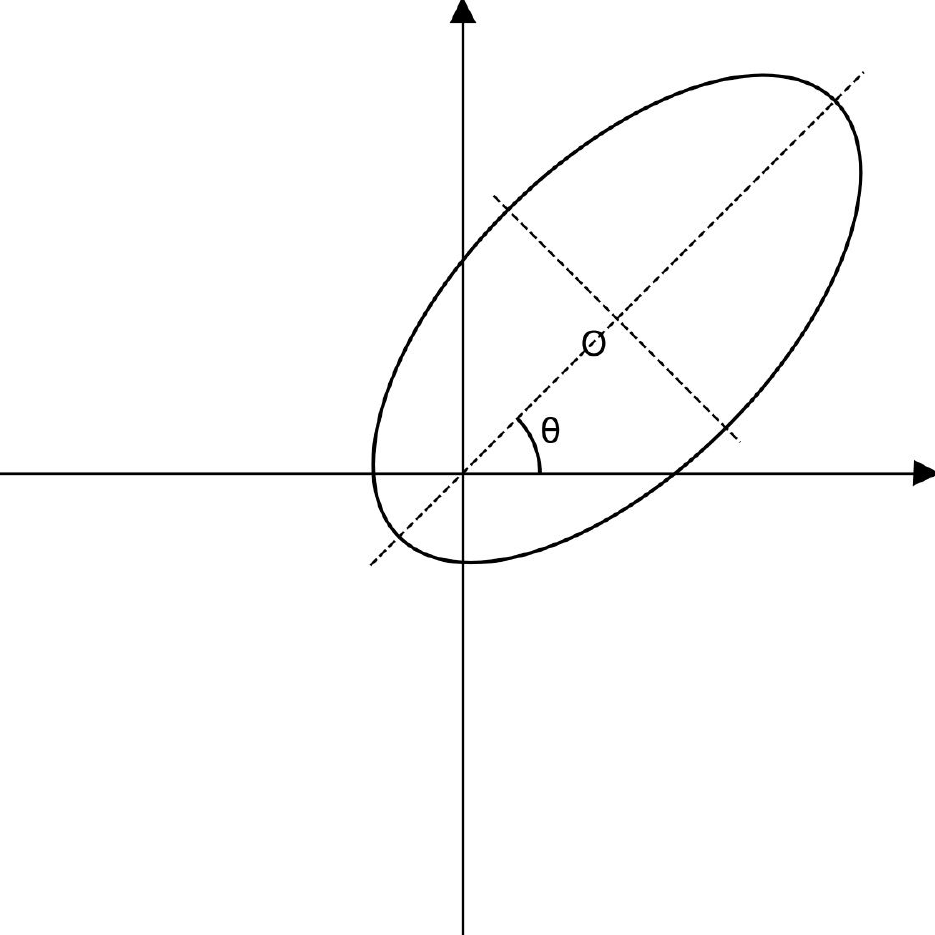} 
        \caption{Blurriness shape with rotation}
    \end{subfigure} \vspace{2mm}%
    ~
    \begin{subfigure}[t]{3.5cm}
        \centering
         \captionsetup{justification=centering}
        \includegraphics[width=3.5cm]{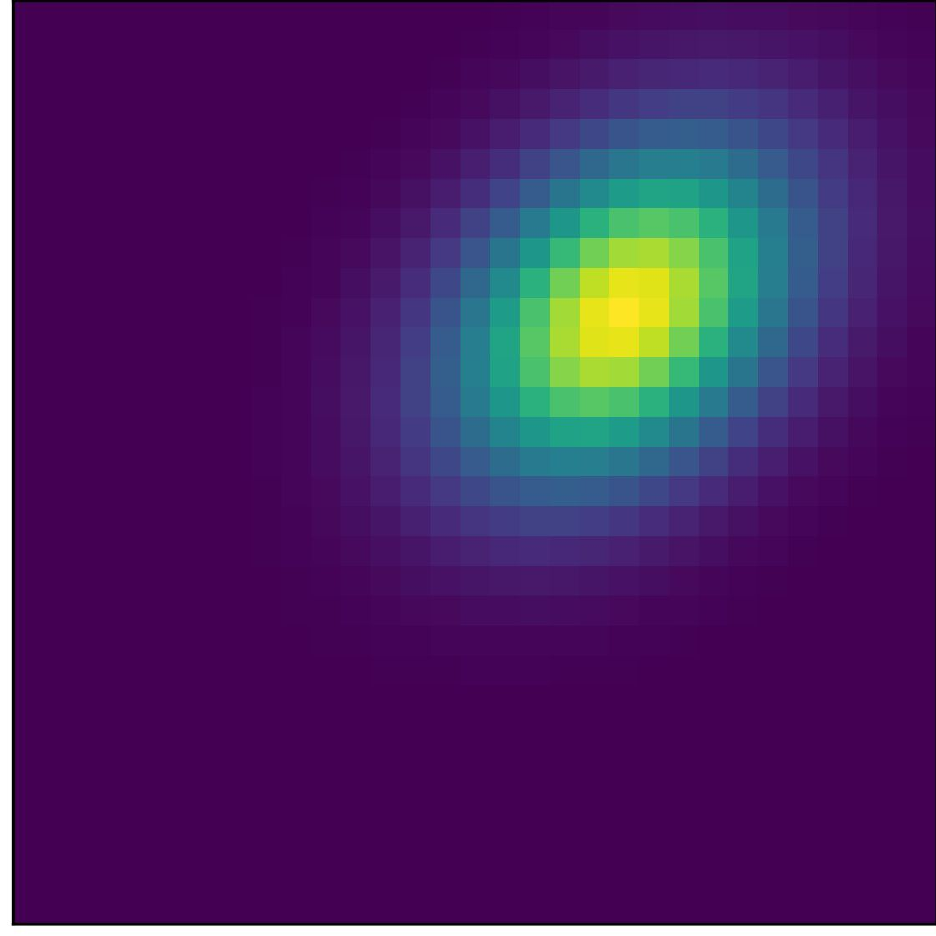}
        \caption{Rotation-enhanced base kernel}
    \end{subfigure} 
        ~
     \begin{subfigure}[t]{3.5cm}
        \centering
         \captionsetup{justification=centering}
        \includegraphics[width=3.5cm]{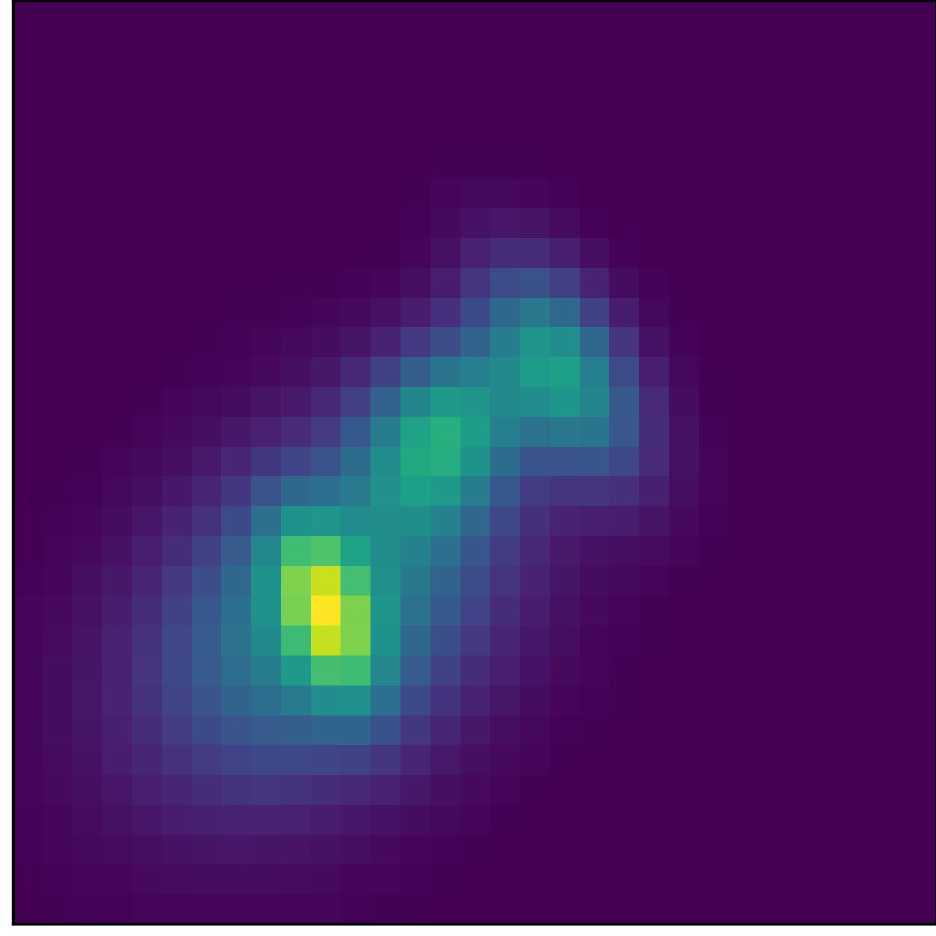}
        \caption{Mixture of $K_{G}^\text{rotation}$}
        \label{fg:finalsp}
    \end{subfigure} 
    \caption{Kernel mixture with rotation enhancement}
    \label{completefigure}
\end{figure}

Still, the blur kernel constructed by the mixture of $K_{G}^\text{scale}$ cannot model asymmetric blurriness or multi-source blurriness because every base kernel is symmetric and centered at zero.  To overcome this weakness, we propose using a non-zero center for each base kernel which is formulated as 
\begin{equation}
\begin{aligned}
    K_{G}^\text{center}(\mathbf{p};\bm{\Sigma}, \bm{\mu}) \propto
    \exp{\left(-\frac{1}{2}
    (\mathbf{p}-\bm{\mu})^T
    \bm{\Sigma}^{-1}
    (\mathbf{p}-\bm{\mu})\right)}
\end{aligned}
\label{non-center gaussian}
\end{equation}
where $\bm{\mu} =  \begin{bmatrix}
    \mu_x & \mu_y
\end{bmatrix}^T$ and $\mu_x$ and $\mu_y$ are the relative locations of the center on each axis. Fig.~\ref{Non-default gaussian} displays the structure of a Gaussian kernel with a non-zero center as well as a mixture of the same kind. With a non-zero center, the $K_{G}^\text{center}$ can be located anywhere within the range where the kernel is defined. By combining multiple of them, the blur kernel as a whole can present an asymmetric shape of blurriness as shown in Fig.~\ref{partc-nondefault}. The mixture is also capable of modeling a sparse blur kernel by combining multiple base kernels with small scales. With this great flexibility, the ultimate shape of the blur kernel will be determined by the underlying shape of blurriness inherent in a blurred image. As such, this blur kernel exhibits almost nonparametric behaviors.

One limitation of the mixture of $K_{G}^\text{center}$ is that the shapes of the base kernels should be parallel to horizontal and vertical axes. We relax this condition and further consider rotations of the base kernels by using a rotation matrix $\mathbf{R}$ and a rotation angle $\theta$:
\begin{equation}
    K_{G}^\text{rotation}(\mathbf{p};\bm{\Sigma}, \bm{\mu}, \theta) \propto
    \exp{\left(-\frac{1}{2}
    (\mathbf{Rp}-\bm{\mu})^T
    \bm{\Sigma}^{-1}
    (\mathbf{Rp}-\bm{\mu})\right)}, 
  \mathbf{R} = \begin{bmatrix}
  \cos \theta & -\sin \theta \\
  \sin \theta & \cos \theta
  \end{bmatrix}.
\label{complete}
\end{equation}
This formulation provides the most enhanced kernel structure, which leverages all the aforementioned structural features. When $\theta=0$, $K_{G}^\text{rotation}$ is equivalent to $K_{G}^\text{center}$. If $\bm{\mu}$ is also a zero vector, the kernel becomes $K_{G}^\text{shape}$. In fact, the shape complexity of each base kernel is determined by a given image if we use the most advanced structure as a base kernel. The mixture of $K_{G}^\text{rotation}$ can model nearly any shape of blurriness thanks to its high flexibility as shown in Fig.~\ref{fg:finalsp}. 

\subsection{Blind Deblurring Formulation}
For blind image deblurring, both the latent image and blur kernel need to be estimated. In a Bayesian framework, the MAP estimates of $\mathbf{K}$ and $\mathbf{I}$ are determined by maximizing the joint posterior distribution, $p(\mathbf{I ,K | B})$, as
\begin{equation}
    \mathbf{I}^*, \mathbf{K}^* = \underset{\mathbf{I, K}}{\mbox{arg\! max}}  \hspace{1mm} p(\mathbf{I ,K | B}) = \underset{\mathbf{I, K}}{\mbox{arg\! max}} \hspace{1mm} p(\mathbf{B | I, K})p(\mathbf{K})p(\mathbf{I})
    \label{MAP-model}
\end{equation}
while assuming independence between $\mathbf{K}$ and $\mathbf{I}$. This MAP estimation requires specifying $p(\mathbf{K})$ and $p(\mathbf{I})$, the prior distributions of the blur kernel and latent image, respectively.

Instead of assuming particular distributions for priors, we apply the alternating optimization technique. To construct an optimization problem, we modify the objective function of the MAP estimation in Eq.~\eqref{MAP-model}.  By taking the negative logarithm, Eq.~\eqref{MAP-model} becomes
\begin{equation}
     \mathbf{I}^*, \mathbf{K}^* = \underset{\mathbf{I, K}}{\mbox{arg\! min}} \hspace{1mm} -log(p(\mathbf{B | I, K})) - log(p(\mathbf{K})) -log(p(\mathbf{I})).
     \label{loglik}
\end{equation}
A more general optimization problem can be formulated by replacing the log probability in Eq.~\eqref{loglik} by some loss functions:
\begin{equation}
     \mathbf{I}^*, \mathbf{K}^* = \underset{\mathbf{I, K}}{\mbox{arg\! min}} \hspace{1mm} \ell (\mathbf{I} * \mathbf{K}, \mathbf{B}) + \lambda_1 \ell_\mathbf{K} (\mathbf{K}) + \lambda_2 \ell_\mathbf{I} (\mathbf{I})
     \label{obj_func}
\end{equation}
where the first term measures the loss occurring from estimating $\mathbf{B}$ as the convolution of the estimates of $\mathbf{I}$ and $\mathbf{K}$, $\ell_\mathbf{K}$ and $\ell_\mathbf{I}$ are prior terms calculating some loss associated with the estimates of $\mathbf{K}$ and $\mathbf{I}$, respectively. $\lambda_1$ and $\lambda_2$  regularizes the prior terms while presenting their relative importance to the first term, the estimation loss.  

Typically, the kernel prior, $\ell_\mathbf{K} (\mathbf{K})$ is used to stabilize the blur kernel, and the image prior, $\ell_\mathbf{I} (\mathbf{I})$, is used to
recover the latent image with sharp edges.  These priors play a significant role in blind deblurring. While this paper focuses on developing a blur kernel with a flexible structure, a proper prior is still needed to control the level of flexibility in the blur kernel. To achieve that, we solve the following minimization problem:
\begin{equation}
    \mathbf{I}^*, \mathbf{K}^* =  \underset{\mathbf{I}, \mathbf{K}}{\mbox{arg\! min}} \hspace{1mm} || \mathbf{I} * \mathbf{K} - \mathbf{B}||_2^2 + \lambda_1||\mathbf{K}||_2^2 
    +\lambda_2 ||\bm{\sigma}^2||_2^2 + \lambda_3(||\nabla \mathbf{I}||_2^2 + ||\mathbf{I}||_2^2)
    \label{Obj_General}
\end{equation}
where $\bm{\sigma}^2= (\sigma_{x,1}^2, \ldots, \sigma_{x,N}^2, \sigma_{y,1}^2, \ldots, \sigma_{y,N}^2)^T$ is a vector including the diagonal elements $(\sigma_{x,t}^2, \sigma_{y,t}^2)$ of the covariance matrix for each base kernel $K_{G, t}$ for $t=1,\ldots,N$, $\nabla I$ is the image gradient, and $||\cdot||_2$ is the $l_2$-norm. We use $||\mathbf{K}||_2^2$ and $ ||\bm{\sigma}^2||_2^2$ as the kernel prior and $||\nabla \mathbf{I}||_2^2+||\mathbf{I}||_2^2$ as the image prior. Regulating $||\mathbf{K}||_2^2$ and $||\mathbf{I}||_2^2$ induces the sparsity of $\mathbf{K}$ and $\mathbf{I}$, respectively. The inclusion of $||\nabla \mathbf{I}||_2^2$ restricts the estimate of $\mathbf{I}$ by eliminating tiny gradient segments but keeping large ones only.  In addition to these common priors, we propose to include $||\bm{\sigma}^2||_2^2$, say a covariance prior, to further regulate the blur kernel.

\begin{algorithm*}[t]
\SetAlgoLined
\KwIn{ Degraded image $\mathbf{B}$, number of base kernels $N$, kernel size $h$}
 Let $i \leftarrow 0$\;
 Initialize the latent image, $\mathbf{I}^0\leftarrow \mathbf{B}$\;
 Generate random numbers to initialize the model parameters, $\bm{\Sigma}_t^0$ and $\bm{\mu}_t^0$ for $t=1,\ldots,N$\;
 Use the model parameters to initialize base kernels $K_{G,t}^0$ for $t = 1,\ldots,N$\;
 Initialize the blur kernel $\mathbf{K}^0$ by combining $K_{G,t}^0$ for $t = 1,\ldots,N$ according to Eq.~\eqref{Kernel-sum}\;
  \Repeat{
$\frac{||\mathbf{K}^{i} - \mathbf{K}^{i-1}||_2}{||\mathbf{K}^{i-1}||_2}<\epsilon \quad \text{and} \quad \frac{||\mathbf{I}^{i} - \mathbf{I}^{i-1}||_2}{||\mathbf{I}^{i-1}||_2}<\epsilon$
}{Update the number of iteration, $i \leftarrow i+1$\;
\textbf{Blur kernel estimation steps:}
\begin{itemize}
    \item[] Given $\mathbf{I}^{i-1}$, estimate $\mathbf{K}^{i}$ by optimizing Eq.~\eqref{Obj-functionkernel} with respect to $\bm{\Sigma}_t^{i}$ and $\bm{\mu}_t^{i}$ for $t=1,\ldots,N$\;
    \item[] Normalize $\mathbf{K}^{i}$\;
\end{itemize}
\textbf{Latent image restoration steps:}
\begin{itemize}
    \item[] Given $\mathbf{K}^{i}$, recover an intermediate latent image $\mathbf{I}^{i}$ by using Eq.~\eqref{FFT-recover_solution}\;
    \item[] Update a tuning parameter, $\lambda_3 \leftarrow \lambda_3/1.1$\;
\end{itemize}
  
 }
  \KwOut{Latent image $\mathbf{I}^{*} \leftarrow \mathbf{I}^{i}$}
 \caption{Alternating optimization for blind deblurring with the kernel mixture}
 \label{algorithm1}
\end{algorithm*}

As done in typical coefficient shrinkage, e.g., ridge regression, the covariance prior enforces insignificant $\sigma_{x,t}^2$ and $\sigma_{y,t}^2$ to be close to zero and makes the corresponding base kernel $K_{G,t}$ almost negligible in modeling blurriness. This property is used to determine the number of base kernels for the kernel mixture.  Instead of predetermining the exact number of base kernels, we include a large enough number of base kernels in the model. Then, some of them with little impact will become negligible due to the covariance prior.

\subsection{Estimation Procedure}
To solve Eq.~\eqref{Obj_General}, we use alternating optimization as described in Alg.~\ref{algorithm1}. The ultimate purpose of Alg.~\ref{algorithm1} is to recover the latent image $\mathbf{I}$ by modeling a proper blur kernel $\mathbf{K}$, given the degraded image $\mathbf{B}$. To construct $\mathbf{K}$ as a kernel mixture, the information about the number of base kernels, $N$, and the kernel size, $h$, is also needed. The kernel size does not need to be the same with the size of $\mathbf{B}$, but it will be sufficient if the blur kernel $\mathbf{K}$ is large enough to model the blurriness in the degraded image $\mathbf{B}$. Once the inputs are ready, we use $\mathbf{B}$ to initialize the latent image, $\mathbf{I}^0$. Then, we generate random numbers to set the initial values of model parameters, initialize the base kernels, and combine them to form the initial blur kernel, $\mathbf{K}^0$. After the initialization, we keep updating the blur kernel and latent image forming an iterative loop. Within a loop, we optimize $\mathbf{K}$ given the latent image at hand, $\mathbf{I}^{i-1}$, and use the resulting optimal solution to update $\mathbf{K}^i$. The updated blur kernel, $\mathbf{K}^i$, is then used to obtain a new estimate of the latent image, $\mathbf{I}^i$.
This procedure will be repeated until the estimates of the latent image and blur kernel converge. In fact, the procedure in the loop consists of two sub-problems, one for blur kernel estimation and another for latent image restoration, which is further elaborated in the following sections.  

\subsubsection{\textbf{Blur kernel}} 
Under the alternating optimization framework, a latent image estimate is given for the blur kernel estimation. Then, the latent image is assumed constant, so the objective function in Eq.~\eqref{Obj_General} reduces to 
\begin{equation}
    E(\mathbf{K}) = || \mathbf{I} * \mathbf{K} - \mathbf{B}||_2^2 + \lambda_1||\mathbf{K}||_2^2
    +\lambda_2 ||\bm{\sigma}^2||_2^2. 
\label{Obj-functionkernel}
\end{equation}
For the blur kernel estimation, we minimize the energy function, $E(\mathbf{K})$, with respect to
the model parameters that compose $\mathbf{K}$, by using the Conjugate Gradient (CG) method. Once a new estimate of $\mathbf{K}$ is obtained, it is normalized to ensure $\sum_{i} \sum_{j} \mathbf{K}_{ij} = 1$.

\subsubsection{\textbf{Latent image}}
The goal of this sub-problem is to recover the latent image given a blur kernel estimate. Based on Eq.~\eqref{Obj_General}, the energy function is formulated as 
\begin{equation}
    E(\mathbf{I}) = || \mathbf{I} * \mathbf{K} - \mathbf{B}||_2^2 + \lambda_3(||\nabla \mathbf{I}||_2^2 + ||\mathbf{I}||_2^2).
\label{Obj-functionimage}
\end{equation}
For the minimization of Eq.~\eqref{Obj-functionimage}, the closed-form solution exists. By using the Fast Fourier Transform (FFT), the solution is formulated \cite{xu2010two} as
\begin{equation}
\small
 \mathbf{I}^{*} = \mathcal{F}^{-1}(\frac{\overline{\mathcal{F}(\mathbf{K})}  \mathcal{F}(\mathbf{B})}{\overline{\mathcal{F}(\mathbf{K})}\mathcal{F}(\mathbf{K}) + \lambda_3[\overline{\mathcal{F}(\bm{\partial_x})}\mathcal{F}(\bm{\partial_x}) + \overline{\mathcal{F}(\bm{\partial_y})}\mathcal{F}(\bm{\partial_y})] + \lambda_3})                        
\label{FFT-recover_solution}
\end{equation}
where $\mathcal{F}(\cdot)$, $\mathcal{F}^{-1}(\cdot)$ are the FFT and inverse FFT operators, respectively, and $\overline{\mathcal{F}(\cdot)}$ denote the complex conjugate operator of FFT. $\bm{\partial_x}$ and $\bm{\partial_y}$ are the horizontal and vertical partial differential operators, respectively. To achieve a better estimate of $\mathbf{I}$, we keep updating $\lambda_3$ at each iteration as has been done in \cite{bai2018graph},\cite{pan2016$l_0$}.

\section{Comparative Experiments}\label{sc:case}
In this section, we use satellite image data and compare the performance of our proposed method with other state-of-the-art methods. We describe the dataset and experimental settings used for the implementation of the methods and discuss the comparison results.

\subsection{Dataset}
The dataset used in this study includes a simulated image of satellite that is convolved with unknown kernels of various noises to generate a wide spectrum of blurred noisy images. These synthetic images are distorted by a two-layer wind model whose strength turbulence parameters create different types of noisy images, characterized as $Dr=10$ (Dr10) and $Dr=20$ (Dr20) \cite{swindle2018high}. The dataset consists of 200 images in total, including 100 images for each distortion type of Dr10 or Dr20. The images are in RGB (three channels) format and their patch size is 365$\times$365. The authors of \citet{swindle2018high} generously provided the 200 blurred images and the original (simulated) satellite image for the analysis of this study. As such, we are not aware of what kernels were used for the convolution and how the final blurry images were created.

\begin{figure}[t]
    \centering
    \begin{subfigure}[h]{4cm}
        \centering
        \includegraphics[width=4cm]{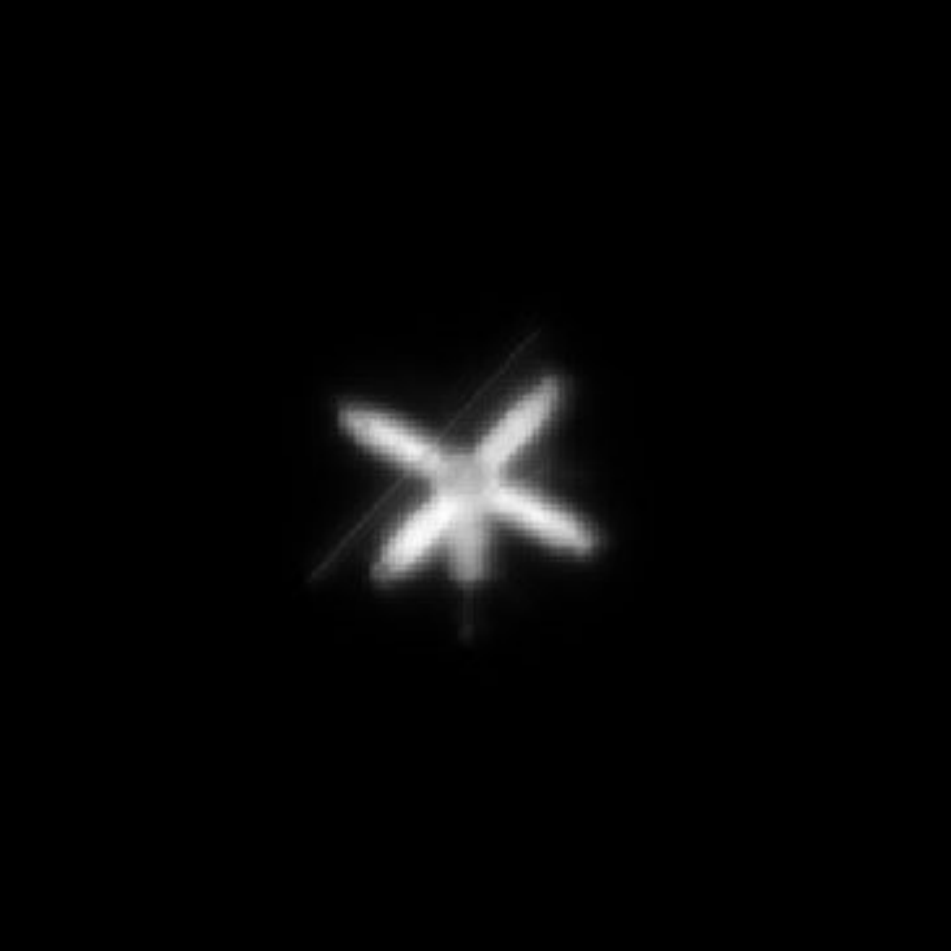}
        \caption{Blurred image at Dr10}
        \label{dr10}
    \end{subfigure} \vspace{2mm}%
    ~
     \begin{subfigure}[h]{4cm}
        \centering
        \includegraphics[width=4cm]{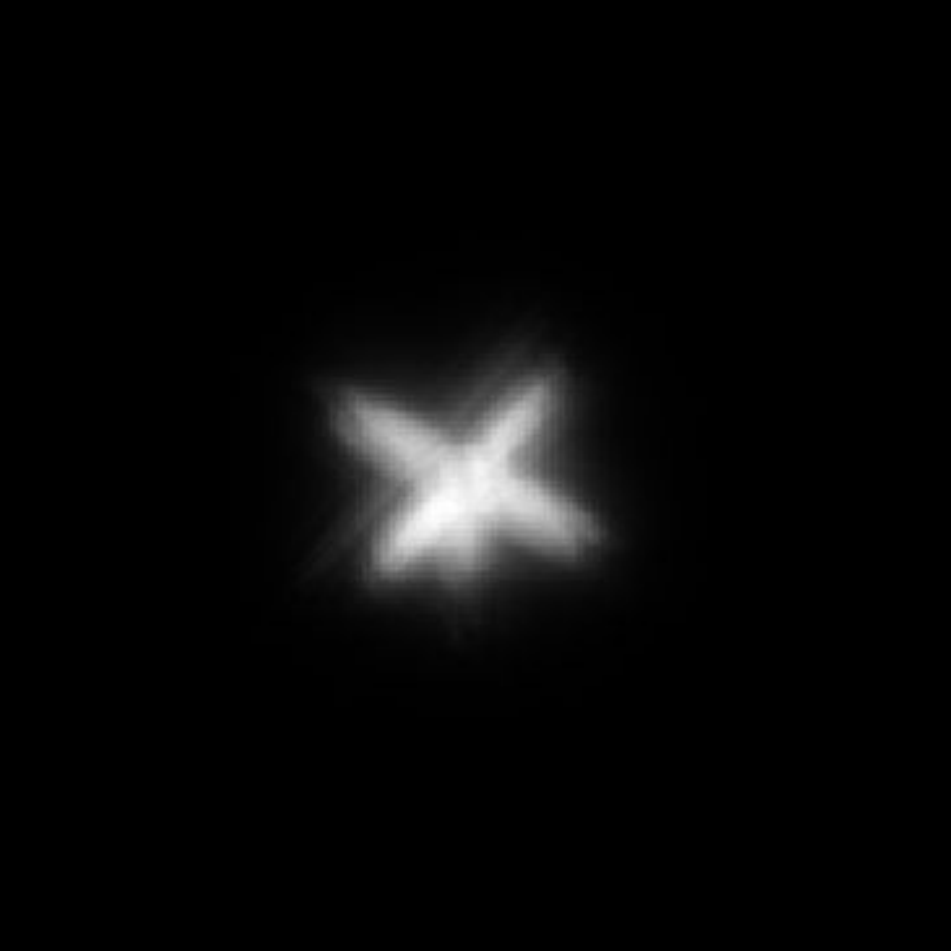}
        \caption{Blurred image at Dr20}
        \label{dr20}
    \end{subfigure}
    ~
    \begin{subfigure}[h]{4cm}
        \centering
        \includegraphics[width=4cm]{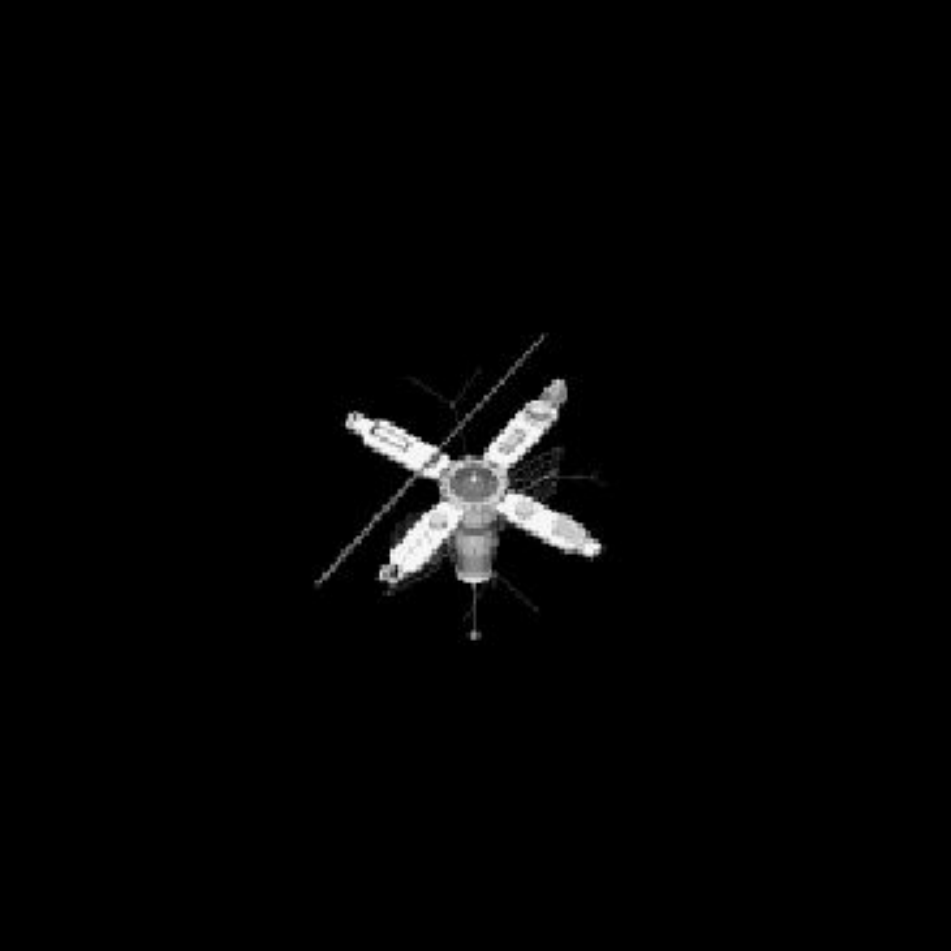}
        \caption{Simulated (real) Image}
        \label{Real Image}
    \end{subfigure}
    \caption{Sample images and the original image without degradation}
    \label{Image-Dr10}
    
\end{figure}

Fig.~\ref{dr10} and \ref{dr20} presents a sample of Dr10 and Dr20 image, respectively. Dr20 images are noisier and more degraded, and it is even hard to distinguish the object of satellite.  Fig.~\ref{Real Image} shows the simulated satellite image without any degradation, which will be referred to as the real image. The different images demonstrate the distortion severity of the dataset, especially for Dr20. 

\subsection{Experimental Settings}
Our proposed method is evaluated in comparison with other state-of-the-art methods in both quantitative and qualitative fashion. The benchmark methods include \citet{xu2010two}, \citet{ren2016image}, \citet{pan2016blind}, and \citet{bai2018graph}.
These papers are not only recent but also have shown their superior performance to others. In addition, their methods are publicly available online as software or Matlab source codes, so the implementation of these methods can be accurate and simple.
\citet{xu2010two} can be implemented via software, but for its implementation, a specific kernel size needs to be supplied.  There are three options of small, medium, and large kernels. The medium kernel provided the best quality of deblurred images when applied \citet{xu2010two} to our data. For a fair comparison, we consider the same size of kernel for all the methods including ours, i.e., $h = 31\times 31$ (the value for the medium size kernel). Our method is implemented in Python on an i7-8700 CPU system.

To fully specify our model and estimation method, we need to determine a set of parameters other than the model parameters. This includes the number of base kernels, $N$, and the regularization parameters of $\lambda_1, \lambda_2$, and $\lambda_3$. 

Fig.~\ref{partc-nondefault} and \ref{fg:finalsp} implies that the number of kernels is one of the most critical factors defining the overall shape of blur kernel. As its impact being so crucial, we do not choose the exact number of kernels with any prior knowledge, but we make the selection adaptive to the given image and underlying blurriness structure therein. This is achieved by modeling the covariance prior in Eq.~\eqref{Obj_General} while leaving the value of $N$ as a sufficiently large number. From our preliminary study, we found that $N=9$ could provide enough level of flexibility for the kernel mixture.
For the regularization parameters, we sampled a few images and selected $\lambda_1=10^{-4}$ and $\lambda_2=10^{-2}$ based on the quantitative metrics described in Section~\ref{ssc:perf} and visual inspection of recovered images. On the other hand, as described in Alg.~\ref{algorithm1}, $\lambda_3$ is kept updated in each iteration of the estimation procedure, which is initialized to $10^{-2}$.

One thing to note is, for the base kernels, we use $K_{G}^\text{center}$ instead of $K_{G}^\text{rotation}$ to form the kernel mixture, $\mathbf{K}$. Although $K_{G}^\text{rotation}$ is the most flexible and capable of modeling the most complicated shape of blurriness, it adds additional $N$ rotation parameters to estimate, one for each base kernel. Then, the total number of parameters to estimate becomes $5N$ which is 45 when $N=9$. From the perspective of the nonlinear optimization in Eq.~\eqref{Obj-functionkernel}, these additional parameters require significantly more computations. At least for the images used in this study, the additional parameters do not add much benefit in terms of the image recovery (also see the similarity between Fig.~\ref{partc-nondefault} and \ref{fg:finalsp}) but increase the variance of the final estimate producing a poorer quality of a recovered image.

\subsection{Performance Measures}\label{ssc:perf}
We use two performance measures to quantify and compare the quality of various blind deblurring methods. The first measure is root mean square error (RMSE) that is widely used 
to evaluate the estimation (or prediction) accuracy of a model not only in the blind deblurring domain but also in general machine learning applications.
The second measure is peak signal-to-noise ratio (PSNR) that is often used in computer vision and image processing applications to quantify the quality of the reconstructed images.

The RMSE and PSNR can be calculated as follows:
\begin{align}
    &\text{RMSE}(\mathbf{I},\hat{\mathbf{I}}) = \sqrt{\frac{1}{n_h n_v}\sum_{i=1}^{n_h} \sum_{j=1}^{n_v} (\mathbf{I}_{ij} - \hat{\mathbf{I}}_{ij})^2}    \label{RMSE}\\
&\text{PSNR}(\mathbf{I},\hat{\mathbf{I}}) =  \frac{\max_{i,j}\hat{\mathbf{I}}_{i,j} - \min_{i,j}\hat{\mathbf{I}}_{i,j}}{\sum_{i=1}^{n_h} \sum_{j=1}^{n_v} (\mathbf{I}_{ij} - \hat{\mathbf{I}}_{ij})^2/n_h n_v}
   \label{PSNR}
\end{align}
where $\mathbf{I}_{ij}$ and $\hat{\mathbf{I}}_{ij}$ for $i=1,\ldots,n_h$ and $j=1,\ldots,n_v$ are the values of $i$th horizontal and $j$th vertical pixel from the real and recovered image, respectively.  Since the size of the images used in this study is $365\times 365$, $n_h=n_v=365$.
The RMSE evaluates how different the real and recovered images are whereas the PSNR quantifies how much (peak) variation is captured by the recovered image relative to the average of remaining variation. As such, the smaller the RMSE is and the larger the PSNR is, the better the quality of a recovered image is. Because they measure different aspects of the quality, we use both of them to compare the proposed method with others.

\subsection{Comparison Results}
We randomly sample 10 images from each of the Dr10 and Dr20 datasets to compare the performance.
\subsubsection{\textbf{Dr10 Results}}
\begin{figure*}[h]
\centering
    \begin{subfigure}[h]{8cm}
        \centering
        \includegraphics[width=8cm]{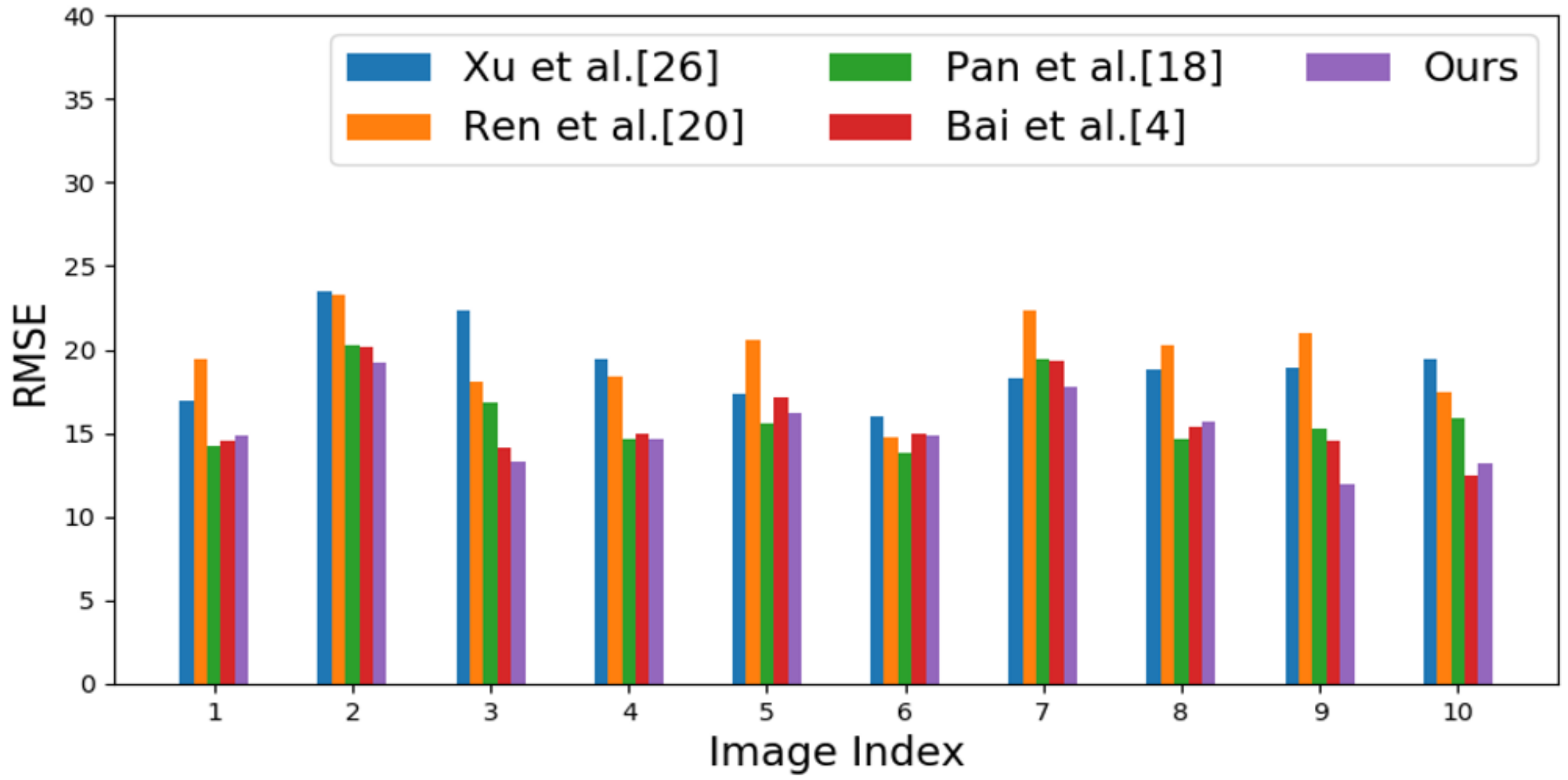} 
        \caption{RMSE results}
    \end{subfigure} \vspace{2mm}%
    ~
     \begin{subfigure}[h]{8cm}
        \centering
        \includegraphics[width=8cm]{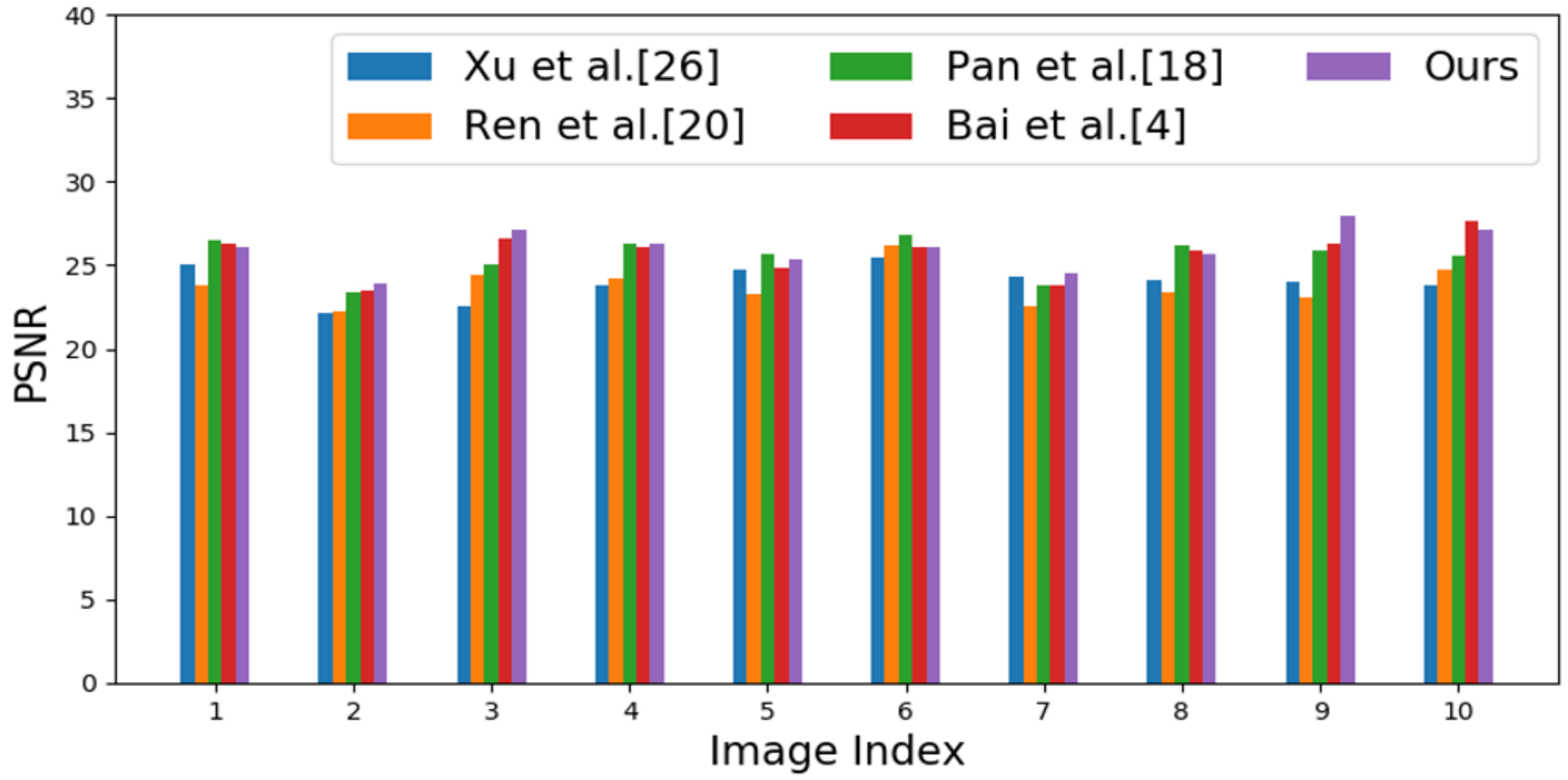}
        \caption{PSNR results}
    \end{subfigure}
    \caption{Visualization of relative performance - Dr10}
    \label{Dr10 measures}
\end{figure*}

Table~\ref{tb:Dr10} shows the RMSE and PSNR values for the ten recovered Dr10 images. In all cases, the proposed method outperforms \citet{xu2010two} and \citet{ren2016image}. In addition, our method performs better in most cases than the most sophisticated method, the graph-based blind deblurring \cite{bai2018graph}. Still, the graph-based method consistently shows descent performance while producing the best recovered image in one case. In couple other cases, the dark channel prior method \cite{pan2016blind} performs the best, but its performance is worse than the graph-based method on average. Even though the dark channel method can provide really great quality for certain images, it can also suffer from considerably poor performance; see the results of Image 3, 9, and 10 where its RMSE and PSNR significantly deviate from the best values. On the contrary, our proposed method not only has the most best cases but also remains close to the best performance whenever it loses the first place. All these observations are applied to both RMSE and PSNR measures. Overall, the proposed method has the lowest RMSE and the highest PSNR on average. Fig.~\ref{Dr10 measures} visualizes the relative performance of all methods, demonstrating the superiority of our method.  

\begin{table*}[b]
\caption{Performance measures for randomly sampled Dr10 images; the boldface highlights the best value for each image.}
\centering
\scriptsize
    \makebox[\textwidth]{\begin{tabular}{|c|c|c|c|c|c||c|c|c|c|c|}
\hline
 \multicolumn{1}{|c|}{} &
 \multicolumn{5}{c||}{\textbf{RMSE}}&
 \multicolumn{5}{c|}{\textbf{PSNR}}\\
 \hline
 Image &   Xu \textit{et al.} \cite{xu2010two} & Ren \textit{et al.} \cite{ren2016image} & Pan \textit{et al.} \cite{pan2016blind} & Bai \textit{et al.} \cite{bai2018graph}  & Ours & Xu \textit{et al.} \cite{xu2010two} & Ren \textit{et al.} \cite{ren2016image} & Pan \textit{et al.} \cite{pan2016blind} & Bai \textit{et al.} \cite{bai2018graph}  & Ours \\
\hline
  1 & 16.89 & 19.47 & \textbf{14.26} & 14.55 & 14.90 & 24.98 & 23.75 & \textbf{26.46} & 26.29 & 26.08\\
  2 & 23.51 & 23.25 & 20.31 & 20.18  & \textbf{19.20} & 22.12 & 22.21 & 23.39 & 23.44  & \textbf{23.87}\\ 
  3 & 22.38 & 18.12 & 16.81 & 14.11  & \textbf{13.27}& 22.54 & 24.38 & 25.03 & 26.55  & \textbf{27.08}\\ 
  4 & 19.43 & 18.44 & 14.62 & 14.95 & \textbf{14.61}&23.77 & 24.22 & 26.24 & 26.05 & \textbf{26.25}\\ 
  5 & 17.33 & 20.54 & \textbf{15.56} & 17.19  &16.20& 24.76 & 23.29 & \textbf{25.70} & 24.84  & 25.35\\
  6  & 16.00 & 14.72 & \textbf{13.77} & 14.96 & 14.84 & 25.45 & 26.18 & \textbf{26.76} & 26.04 & 26.11 \\ 
  7 & 18.29 & 22.33 & 19.40 & 19.27  & \textbf{17.73}& 24.30 & 22.56 & 23.78 & 23.84  & \textbf{24.57}\\ 
  8 & 18.78 & 20.29 & \textbf{14.68} & 15.34  & 15.71 & 24.10 & 23.40 & \textbf{26.21} & 25.83  & 25.62\\
  9 & 18.90 & 20.99 & 15.27 & 14.50  & \textbf{11.96}& 24.01 & 23.10 & 25.86 & 26.31  & \textbf{27.99}\\ 
  10 & 19.41 & 13.50 & 15.88 & \textbf{12.49}  & 13.19 & 23.78 & 24.72 & 25.52 & \textbf{27.61}  & 27.14\\ \hline

  Mean & 19.10 & 19.16 & 16.05 & 15.75  & \textbf{15.16} &23.98 & 23.78 & 25.50 & 25.68  & \textbf{26.00}\\

 \hline
\end{tabular}}

\label{tb:Dr10}
\end{table*}

\begin{figure}[h]
    \centering
    \begin{subfigure}[h]{5cm}
        \centering
        \includegraphics[width=4.5cm]{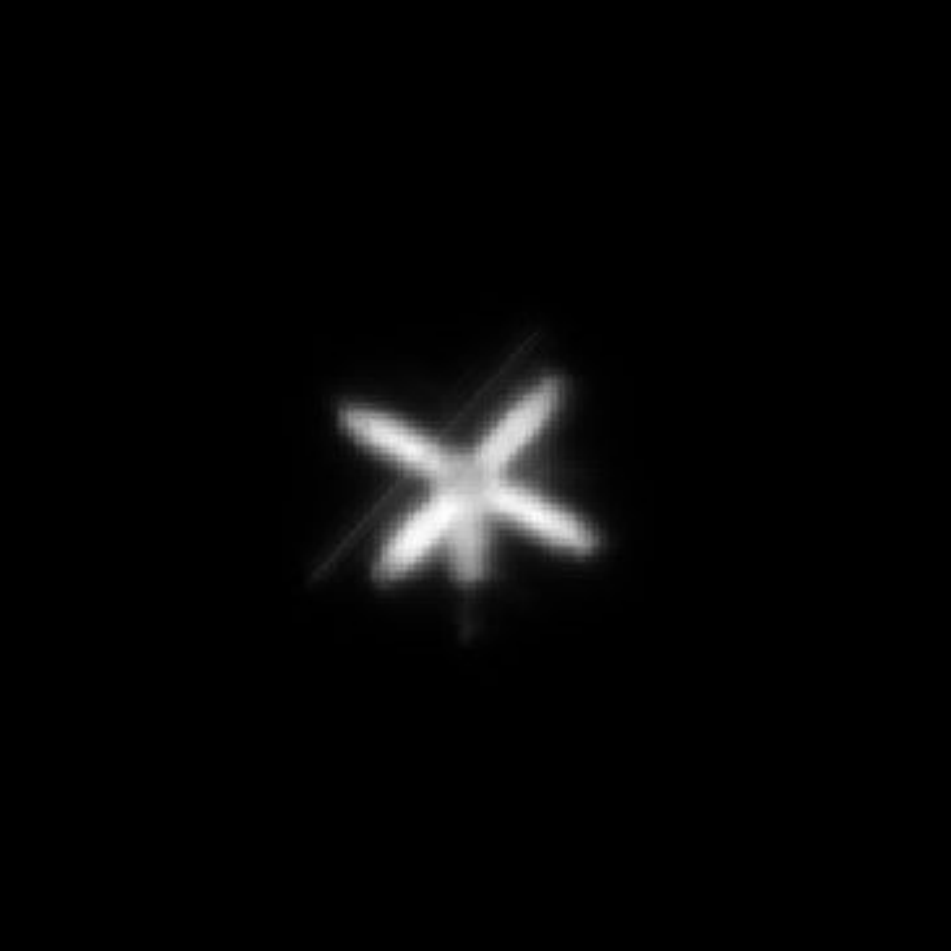}
        \caption{}
    \end{subfigure} \vspace{2mm}%
    ~
    \begin{subfigure}[h]{5cm}
        \centering
        \includegraphics[width=4.5cm]{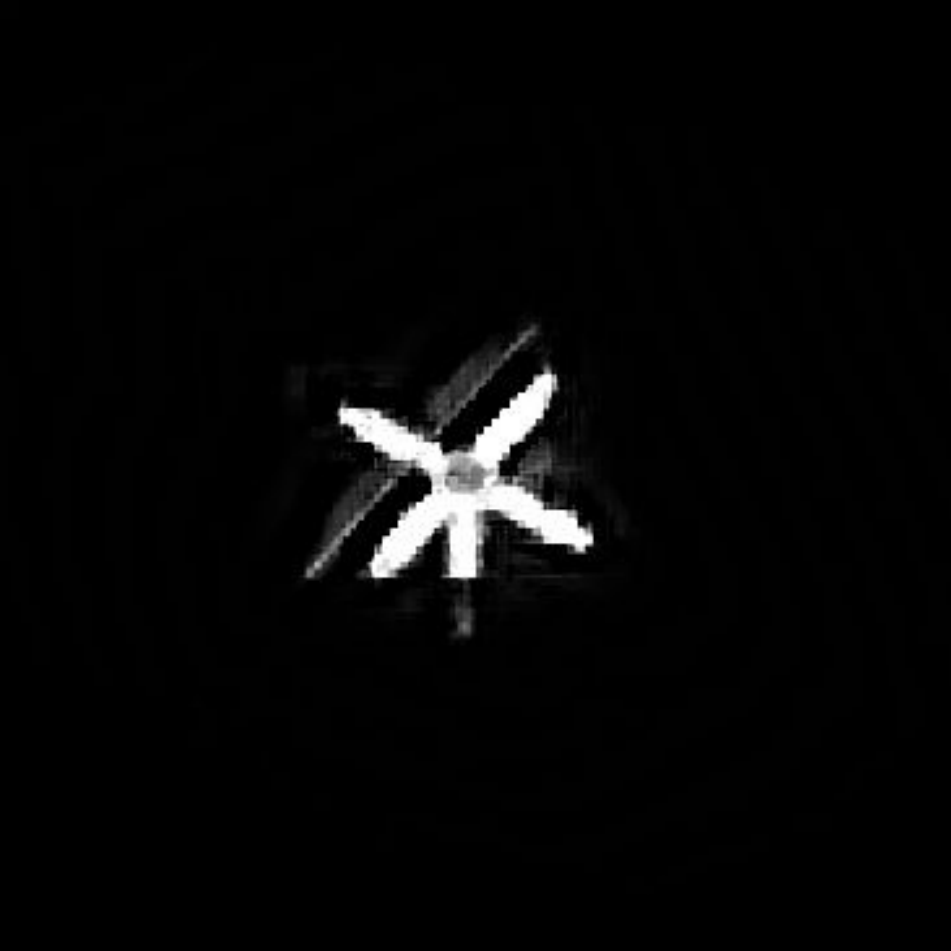}
        \caption{}
         \label{software sample1}
    \end{subfigure}
    ~
     \begin{subfigure}[h]{5cm}
        \centering
        \includegraphics[width=4.5cm]{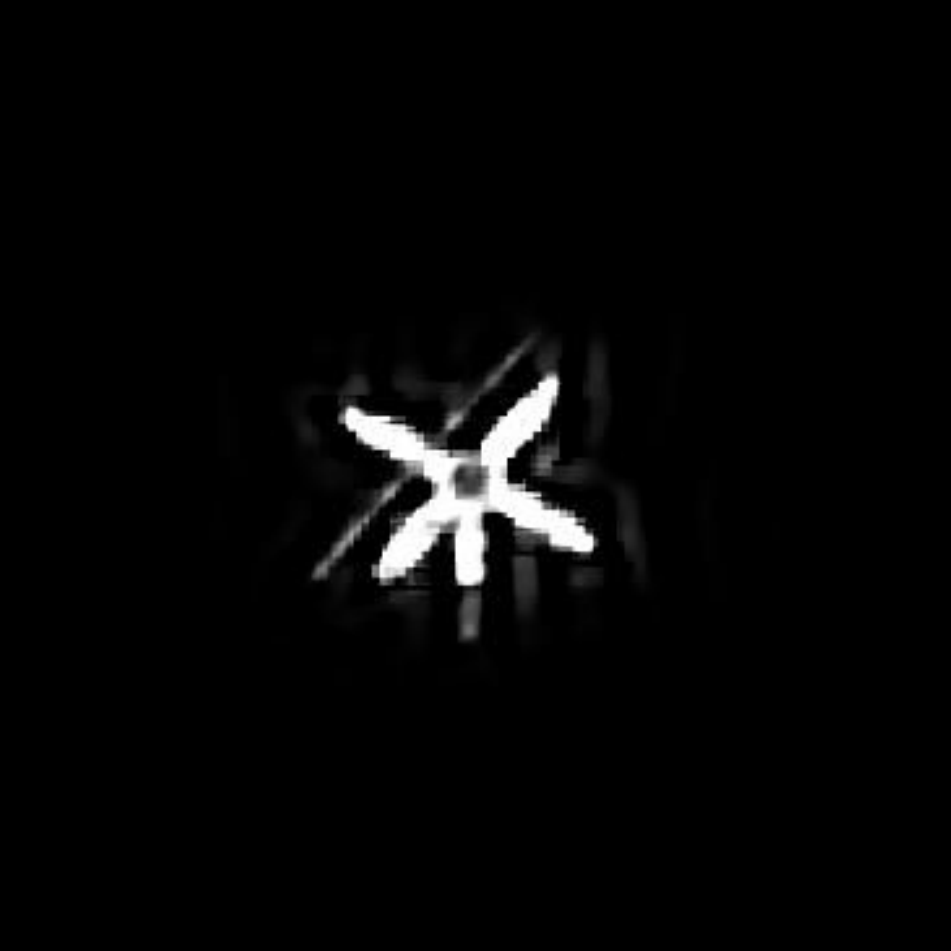}
        \caption{}
        \label{rank sample1}
    \end{subfigure}
    ~
    \begin{subfigure}[h]{5cm}
        \centering
        \includegraphics[width=4.5cm]{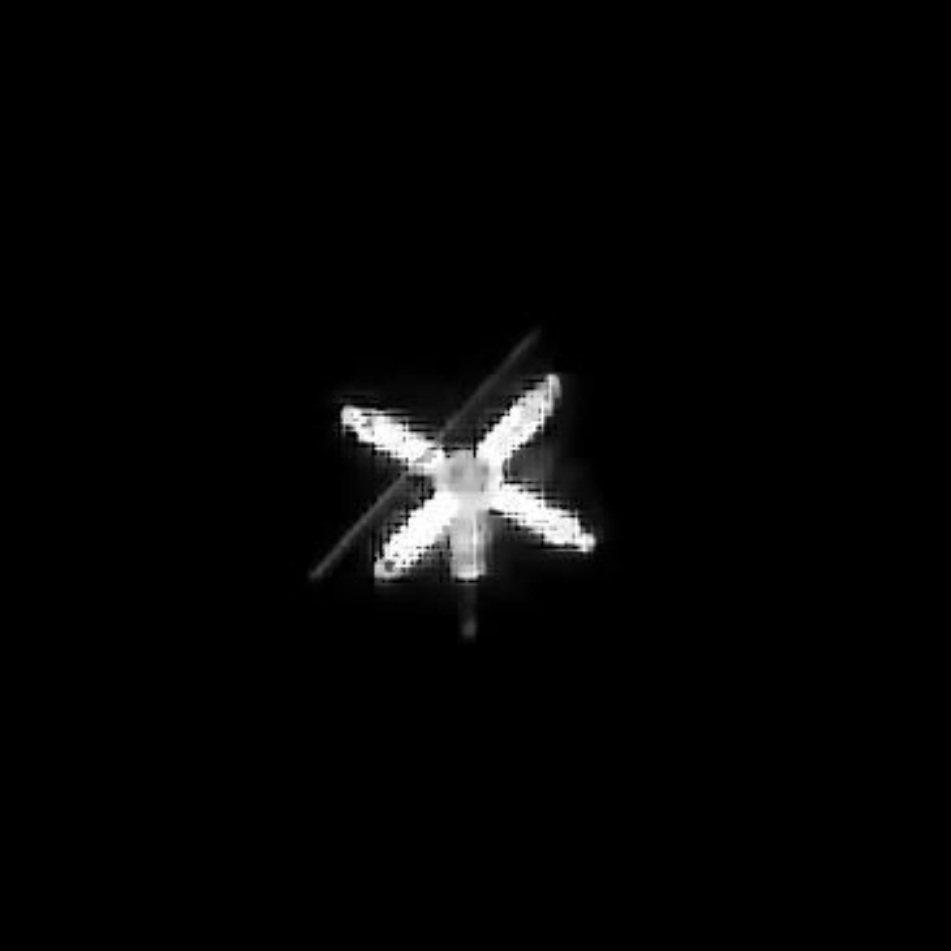}
        \caption{}
        \label{dark sample1}
    \end{subfigure}
    ~
     \begin{subfigure}[h]{5cm}
        \centering
        \includegraphics[width=4.5cm]{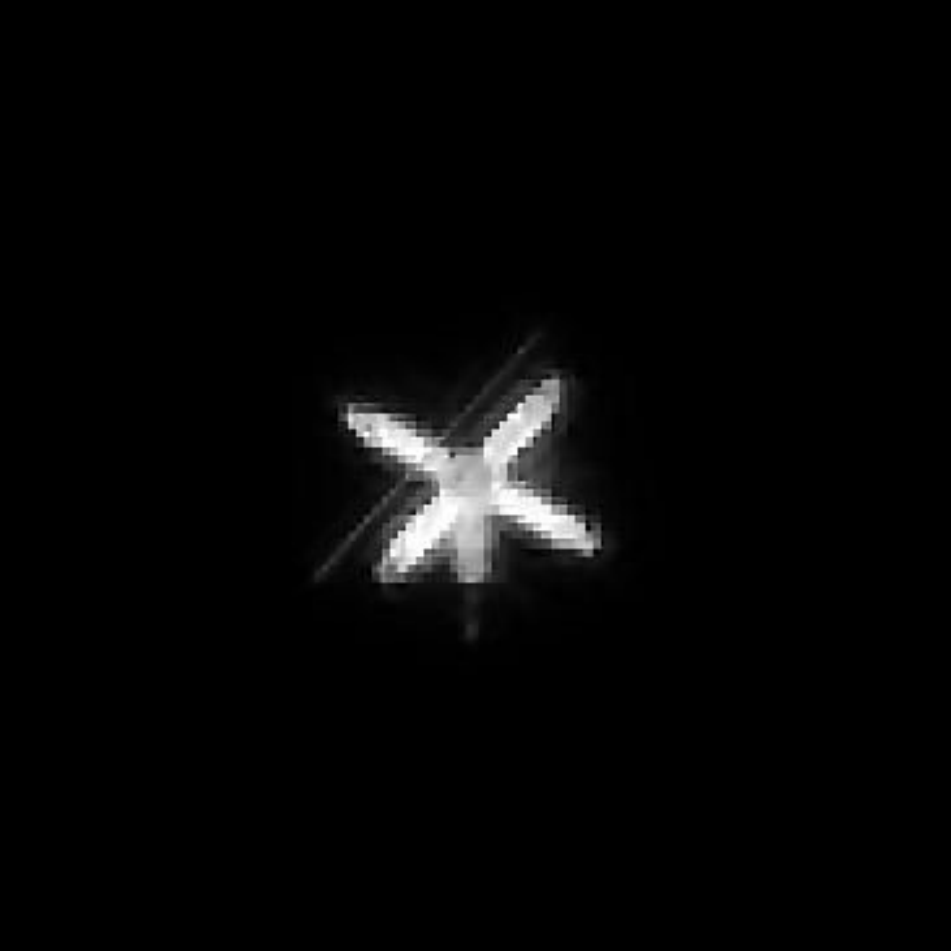}
        \caption{}
        \label{graph sample1}
    \end{subfigure}
    ~
    \begin{subfigure}[h]{5cm}
        \centering
        \includegraphics[width=4.5cm]{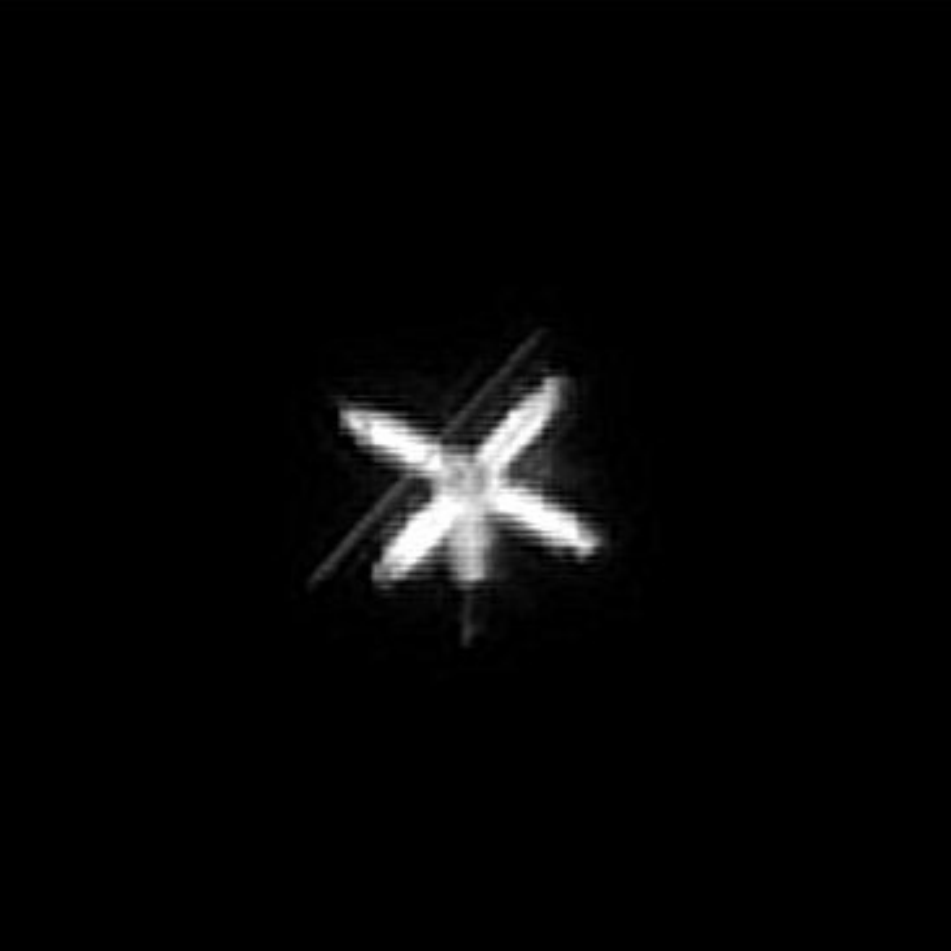}
        \caption{}
         \label{ours sample1}
    \end{subfigure}
    \caption{Dr10 - Results output (Image 3): (a) Blurred images, (b) \citet{xu2010two}, (c) \citet{ren2016image}, (d) \citet{pan2016blind}, (e) \citet{bai2018graph}, (f) Proposed method}
    \label{Image-Dr10-1}
\end{figure}

\begin{figure}[t]
    \centering
    \begin{subfigure}[h]{5cm}
        \centering
        \includegraphics[width=4.5cm]{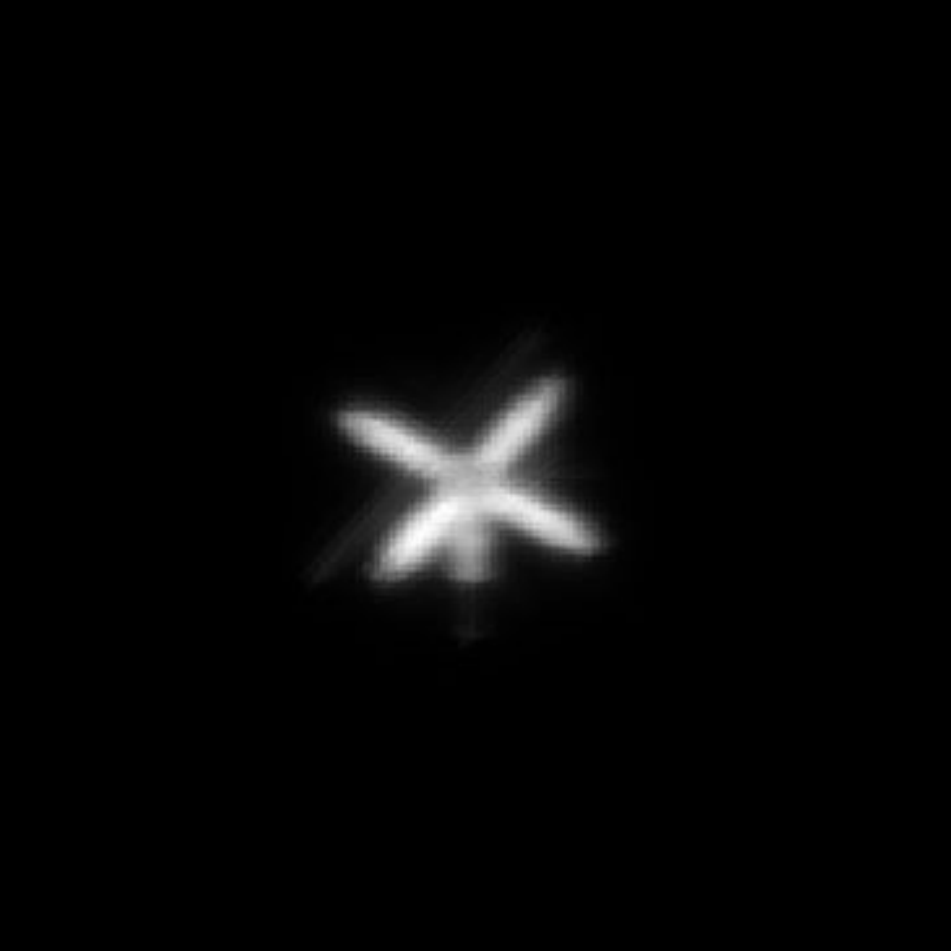}
        \caption{}
    \end{subfigure} \vspace{2mm}%
    ~
    \begin{subfigure}[h]{5cm}
        \centering
        \includegraphics[width=4.5cm]{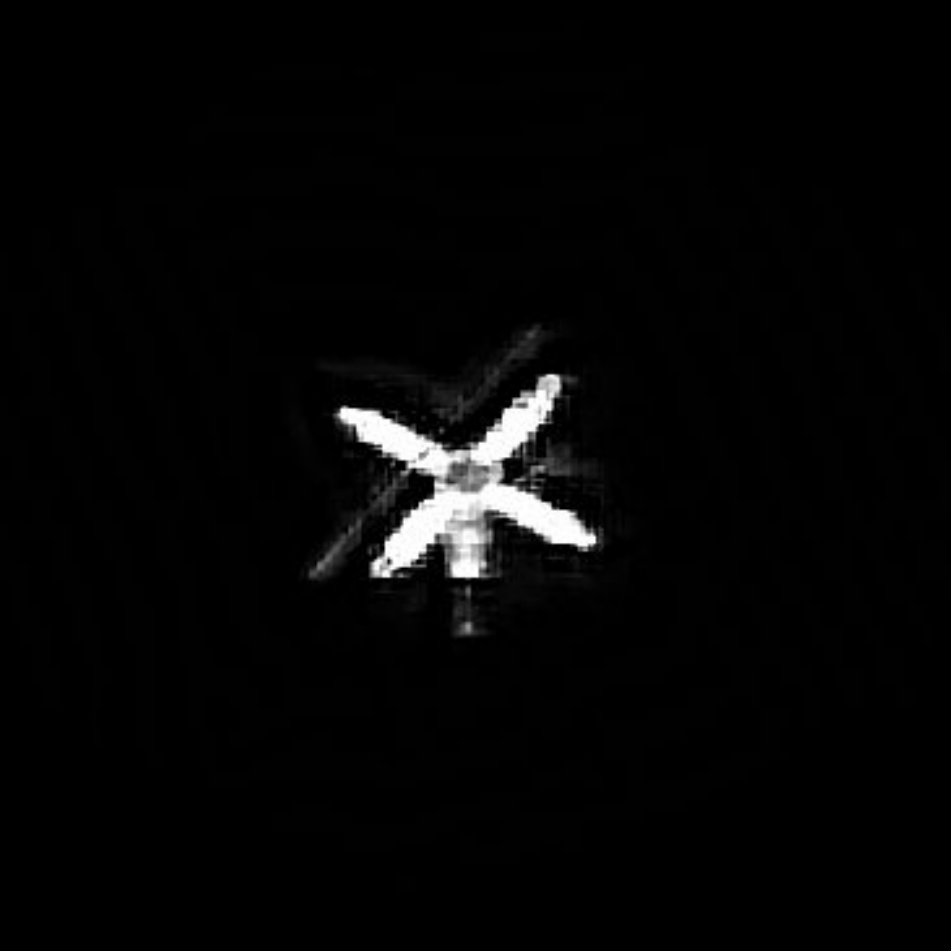}
        \caption{}
        \label{software sample2}
    \end{subfigure}
    ~
     \begin{subfigure}[h]{5cm}
        \centering
        \includegraphics[width=4.5cm]{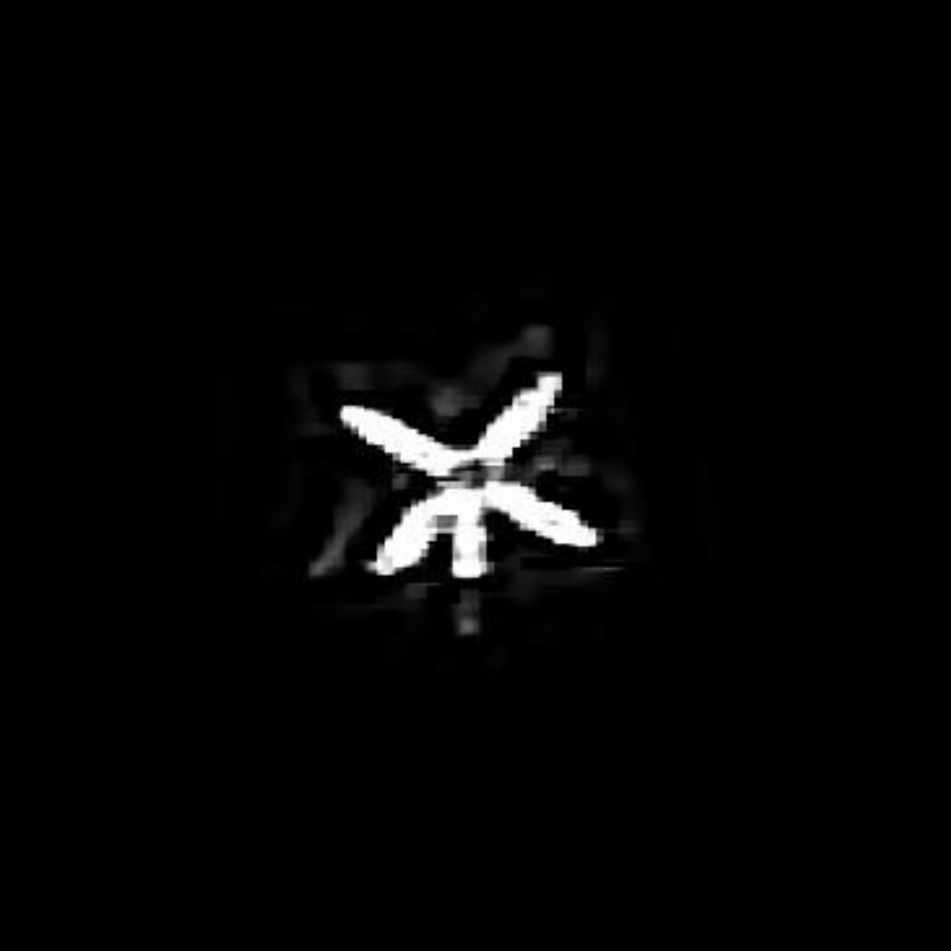}
        \caption{}
        \label{rank sample2}
    \end{subfigure}
    ~
    \begin{subfigure}[h]{5cm}
        \centering
        \includegraphics[width=4.5cm]{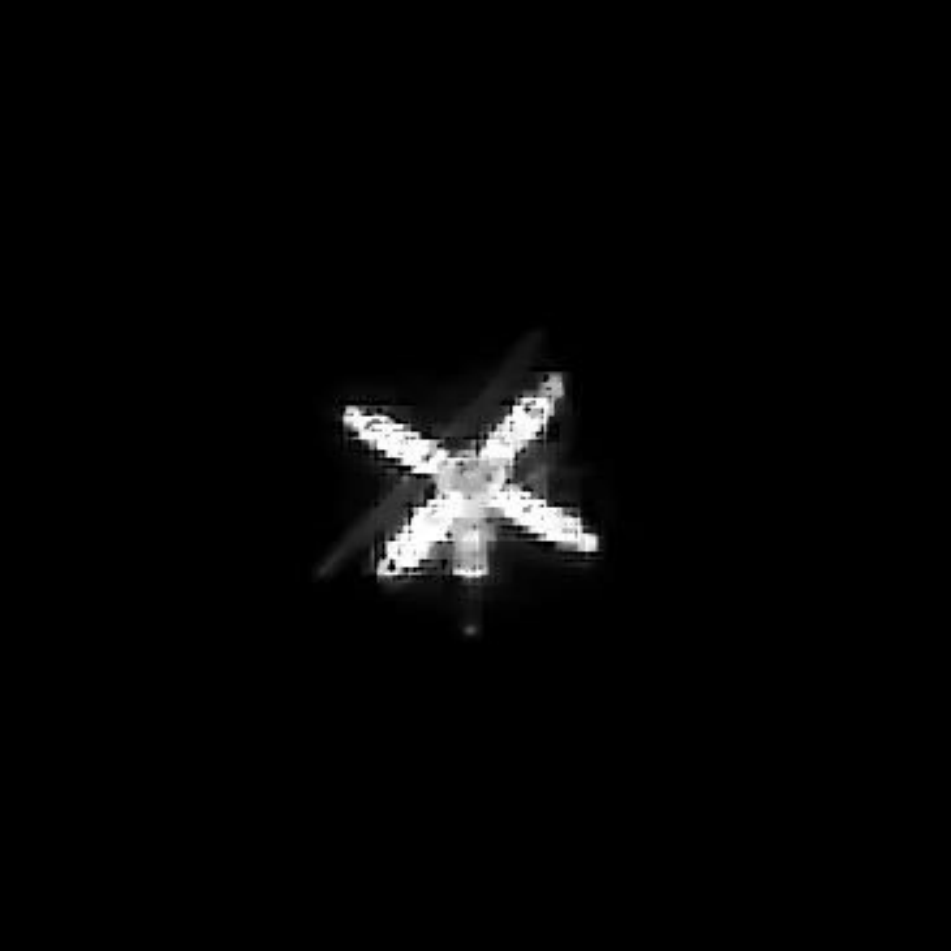}
        \caption{}
        \label{dark sample2}
    \end{subfigure}
    ~
    \begin{subfigure}[h]{5cm}
        \centering
        \includegraphics[width=4.5cm]{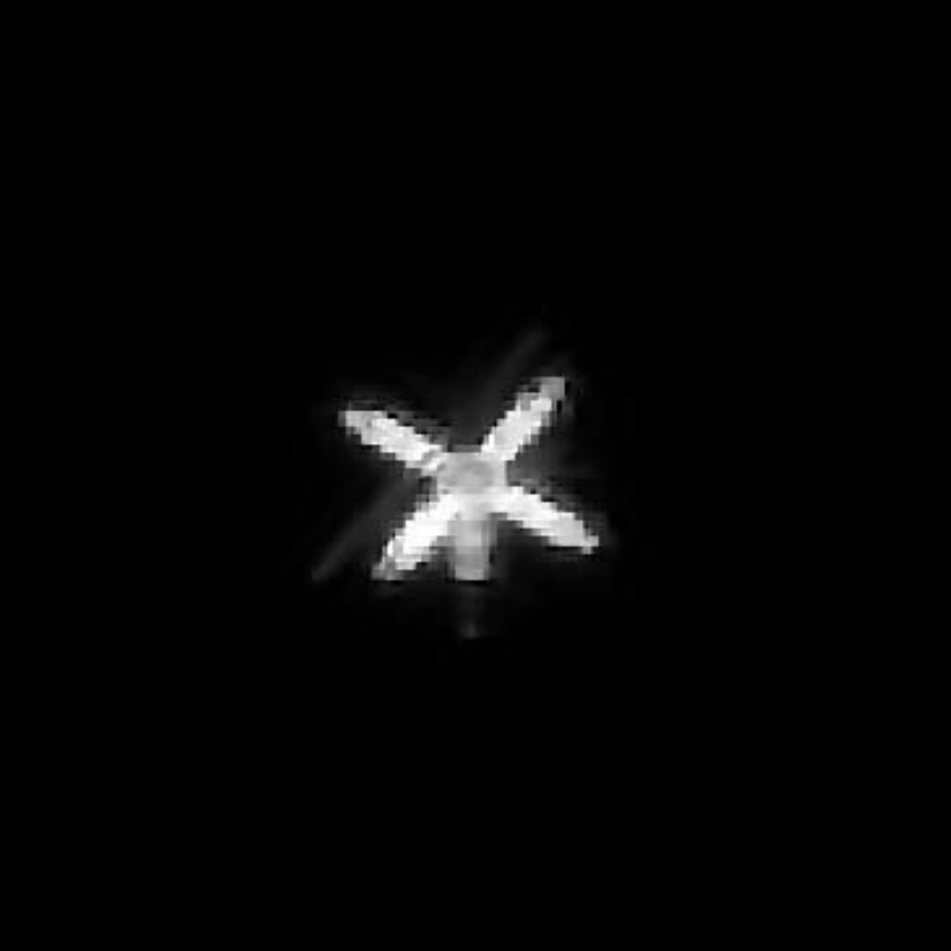}
        \caption{}
        \label{graph sample2}
    \end{subfigure}
    ~
    \begin{subfigure}[h]{5cm}
        \centering
        \includegraphics[width=4.5cm]{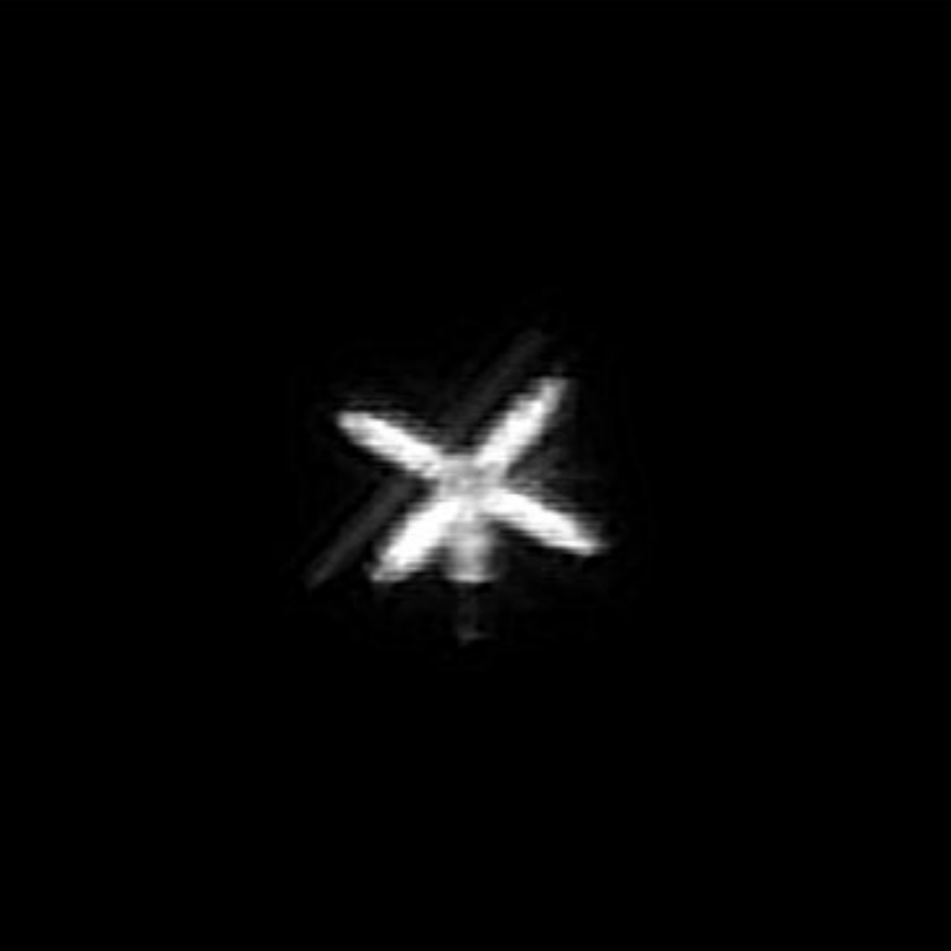}
        \caption{}
        
    \end{subfigure}
    \caption{Dr10 - Results output (Image 9): (a) Blurred images, (b) \citet{xu2010two}, (c) \citet{ren2016image}, (d) \citet{pan2016blind}, (e) \citet{bai2018graph}, (f) Proposed method}
    \label{Image-Dr10-2}
\end{figure}

Fig.~\ref{Image-Dr10-1} and \ref{Image-Dr10-2} illustrate some deblurring outcomes from all methods as well as the blurred image that has been used as an input (Image 3 and 9, respectively). As expected, the visual outcomes are in accordance with the results derived from the quantitative measures. By taking a closer look into the deblurred images, we observe that the dark channel method (Fig.~\ref{dark sample1}) and the graph-based method (Fig.~\ref{graph sample1}) create non-smooth edges of the object, especially around the wings of the satellite. On the contrary, the recovered image from our method (Fig.~\ref{ours sample1}) shows fine edges around the object. The difference in the continuity (or smoothness) of nearby pixel values is reflected in the RMSE and PSNR calculation where our method attains better values than the other two. Fig.~\ref{software sample1} and \ref{rank sample1} show crude reconstruction of the images without much details on the surface of the satellite, which causes higher RMSE and lower PSNR for the corresponding methods.

Fig.~\ref{dark sample1} shows a higher contrast than Fig.~\ref{graph sample1} and \ref{ours sample1} though its contrast level is still lower than that of Fig.~\ref{software sample1} and \ref{rank sample1}. This is likely because the dark channel method regularizes the dark channel of the image producing more of darker pixels. By having a relatively high contrast, the image in Fig.~\ref{dark sample1} looks clearer at distance. However, the high contrast, when it is not appropriate, can result in poorer performance in the image recovery (see Table~\ref{tb:Dr10} for Image 3 and 9).

The similar argument can be made for Fig.~\ref{Image-Dr10-2}. The recovered images from \citet{xu2010two} and \citet{ren2016image} lack detail. The dark channel method and the graph-based method generate even more corrupted edges of the object. The image recovered from the dark channel method has a higher contrast than the images from the graph-based method and our method.
All the results in terms of RMSE, PSNR, and visual inspection demonstrate that the proposed method is competitive with and superior to other state-of-the-art methods in recovering the latent images.

\begin{table*}[b]
\caption{Performance measures for randomly sampled Dr20 images; the boldface highlights the best value for each image.}
\centering
\scriptsize
    \makebox[\textwidth]{\begin{tabular}{|c|c|c|c|c|c||c|c|c|c|c|}
\hline

 \multicolumn{1}{|c|}{} &
 \multicolumn{5}{c||}{\textbf{RMSE}}&
 \multicolumn{5}{c|}{\textbf{PSNR}}\\
 \hline
 Image &   Xu \textit{et al.} \cite{xu2010two} & Ren \textit{et al.} \cite{ren2016image} & Pan \textit{et al.} \cite{pan2016blind} & Bai \textit{et al.} \cite{bai2018graph}  & Ours & Xu \textit{et al.} \cite{xu2010two} & Ren \textit{et al.} \cite{ren2016image} & Pan \textit{et al.} \cite{pan2016blind} & Bai \textit{et al.} \cite{bai2018graph}  & Ours \\
\hline
1  & 23.42 & 23.80 & 21.67 & 22.64 & \textbf{21.08} & 22.15 & 22.01 & 22.82 & 22.44  & \textbf{23.06}\\
  2  & 24.80 & 26.35 & 26.17 & 24.23  & \textbf{23.34} & 21.65 & 21.13 & 21.18 & 21.85  & \textbf{22.18}\\
  3  & 24.02 & 25.49 & 23.45 & 20.57  & \textbf{20.37} & 21.93 & 21.42 & 22.14 & 23.28  & \textbf{23.36}\\ 
  4  & 23.28 & 26.19 & 24.18 & 22.78 & \textbf{21.18} & 22.20 & 21.18 & 21.87 & 22.39  & \textbf{23.02}\\ 
  5  & 21.06 & 24.32 & 22.56 & 22.14  & \textbf{18.84} & 23.07 & 21.82 & 22.47 & 22.64  & \textbf{24.04}\\
  6   & 23.85 & 21.93 & 23.69 & 23.29 & \textbf{21.26} & 21.99 & 22.72 & 22.05 & 22.20  & \textbf{22.99}\\ 
  7  & 23.42 & 25.05 & 24.56 & 24.09  & \textbf{22.61} & 22.15 & 21.57 & 21.74 & 21.90  & \textbf{22.45}\\ 
  8  & 22.60 & 21.80 & 19.97 & 21.24  & \textbf{19.90} & 22.46 & 22.77 & 23.53 & 23.00  & \textbf{23.56}\\
  9  & 19.75 & 22.59 & 20.28 & \textbf{19.77}  & \textbf{19.77} & 23.63 & 22.46 & 23.40 & \textbf{23.62}  & \textbf{23.62}\\ 
  10  & 20.47 & 21.92 & 19.39 & \textbf{17.96} & 18.17 & 23.32 & 22.73 & 23.79 & \textbf{24.45}  & 24.35\\ \hline

  Mean & 22.67 & 23.94 & 22.59 & 21.87  & \textbf{20.65} & 22.45 & 21.98 & 22.49 &  22.77 & \textbf{23.26}\\

 \hline
\end{tabular}}

\label{Dr20-PSNR}
\end{table*}

\begin{figure*}[t]
\centering
    \begin{subfigure}[h]{8 cm}
        \centering
        \includegraphics[width=8cm]{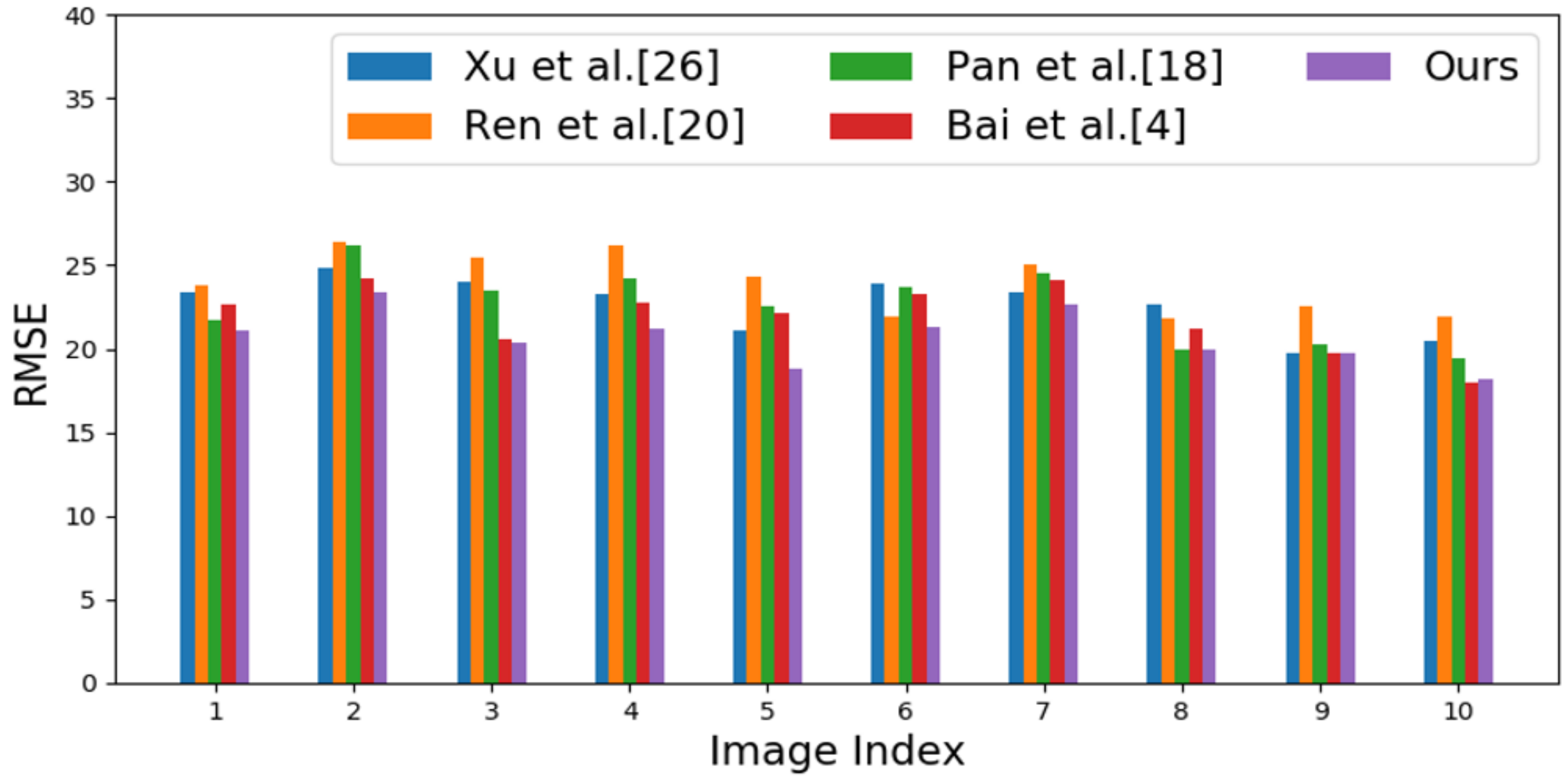} 
        \caption{RMSE results}
    \end{subfigure} \vspace{2mm}%
    ~
     \begin{subfigure}[h]{8cm}
        \centering
        \includegraphics[width=8cm]{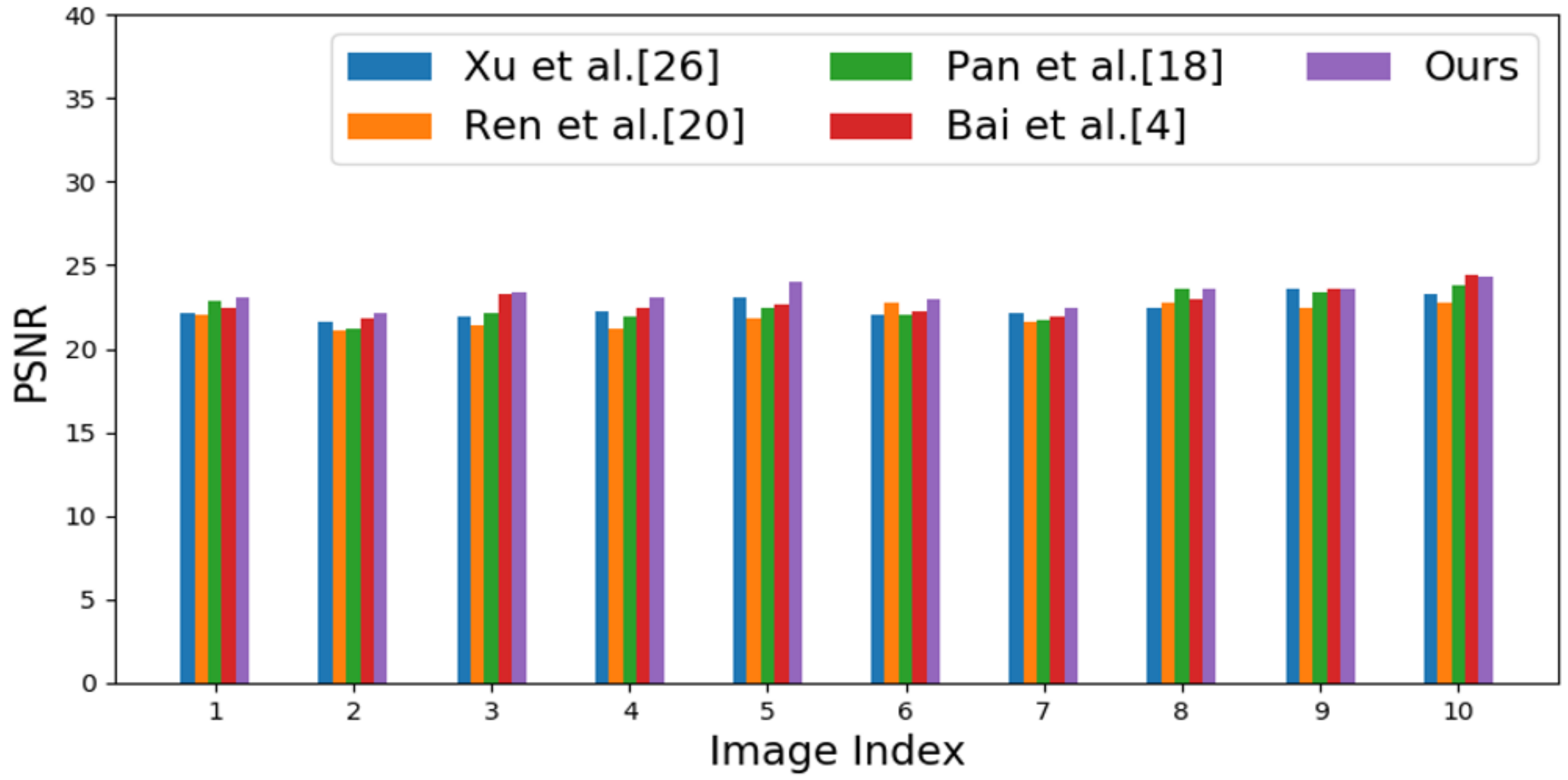}
        \caption{PSNR results}
    \end{subfigure}
    \caption{Visualization of relative performance - Dr20}
    \label{Dr20 measures}
\end{figure*}

\subsubsection{\textbf{Dr20 Results}}

As discussed earlier, Dr20 images involve more blurriness, so it is natural to observe worse performance of the methods compared to Dr10 results.
Table~\ref{Dr20-PSNR} shows the RMSE and PSNR calculations for randomly chosen ten Dr20 images. For these images with a higher level of noises, the results show the overwhelming superiority of the proposed method. taking advantage of explicitly modeling the structure of the blur kernel. Our method attains the lowest RMSE and the highest PSNR in all but one cases. Even for the case where the graph-based method achieves the best quality, our method is the only method producing a competitive result to the best value. This superior performance is likely because our method explicitly models the structure of the blur kernel that represents blurriness.

\section{Concluding Remarks}\label{sc:concl}
In this paper, we propose a novel blind image deblurring method that imposes a specific structure on the blur kernel and achieves great flexibility in modeling blurriness. 
To this end, we develop structure-enhanced Gaussian kernels and form a mixture of the kernels to model the blur kernel. While the behavior of the resulting blur kernel is regulated within a parametric structure, it can represent various shapes of blurriness. Still, the modeling capability and flexibility of the blur kernel depend on 
how many kernels to incorporate in the kernel mixture.
To address this issue, we reformulate the optimization framework for kernel estimation and let the optimization process itself decide
the number of kernels through a covariance prior of the blur kernel.

Our experimental results based on satellite image data show that the proposed method
outperforms descent state-of-the-art methods
that employed some complex image priors, in both quantitative and qualitative manners.
Our method attains the lowest RMSE and the highest PSNR, and this superiority becomes more apparent when the noise level gets higher. From visual inspection, while other methods suffer from discontinuity in nearby pixel values, the proposed method recovers images with smooth edges without any corruption. Yet, the purpose of this method regularizing the blur kernel is not in replacing image prior-based methods but to advance general blind deconvolution practices through a proper combination with the image prior-based methods.

In this paper, we combined several kernels in an additive way which was the simplest form of integrating kernels. Applying a more advanced method for the kernel fusion such as Gaussian Conditional Random Fields would be an interesting research topic. In addition, by improving the optimization process of kernel estimation, the proposed method can be applied to much broader deblurring problems of significant importance. Therefore, finding better heuristics and deriving a closed form solution based on proper assumptions would be a valuable research task. 

\section*{Acknowledgment}
The authors would like to thank Ryan Swindle, Douglas Hope, Michael Hart, and Stuart Jefferies for providing the dataset used in this study.


\bibliography{references} 
\bibliographystyle{plainnat}

\end{document}